\documentclass[sigconf]{acmart}

\renewcommand\footnotetextcopyrightpermission[1]{} 
\usepackage{xspace}
\usepackage{graphicx} 
\usepackage{algorithmic}
\usepackage{mathrsfs}
\usepackage{bm}
\usepackage{amsmath}

\usepackage{amssymb}
\usepackage{booktabs}
\usepackage{pifont}
\usepackage{multirow}
\usepackage{hyperref}
\usepackage{cancel}
\usepackage{dsfont}
\usepackage{subfig}
\usepackage{bbm}
\usepackage{marvosym}

\usepackage[ruled,vlined]{algorithm2e}

\newcommand{\etal}{\textit{et al}.}
\newcommand{\etc}{\textit{etc}.}
\newcommand{\ie}{\textit{i}.\textit{e}.}
\newcommand{\eg}{\textit{e}.\textit{g}.}
\newcommand{\ourmethod}{\textsc{HHCH}\xspace}

\definecolor{linkcolor}{RGB}{218, 50, 138}

\AtBeginDocument{%
  }

\setcopyright{acmcopyright}
\copyrightyear{2023}
\acmYear{2023}
\acmDOI{XXXXXXX.XXXXXXX}

\acmConference[Conference acronym 'XX]{Make sure to enter the correct
  conference title from your rights confirmation emai}{June 03--05,
  2018}{Woodstock, NY}
\acmPrice{15.00}
\acmISBN{978-1-4503-XXXX-X/18/06}

\acmSubmissionID{5108}

\begin{document}

\title{Hyperbolic Hierarchical Contrastive Hashing}

\author{Rukai Wei$^{1}$, Yu Liu$^1$$*$\thanks{$*$ corresponding author}, Jingkuan Song$^2$, Yanzhao Xie$^1$, Ke Zhou$^1$}
\affiliation{
$^1$Huazhong University of Science and Technology\\$^2$University of Electronic Science and Technology of China\country{}}
\affiliation{
$^1$\{weirukai, yu\_liu,  yzxie, zhke\}@hust.edu.cn \\$^2$\{jingkuan.song\}@gmail.com\country{}
}

\renewcommand{\shortauthors}{RK wei, et al.}
\renewcommand{\authors}{RK wei, Y Liu, JK Song, YZ Xie, K Zhou}
\renewcommand{\shortauthors}{Rukai Wei, et al.}

\begin{abstract}
Hierarchical semantic structures, naturally existing in real-world datasets, can assist in capturing the latent distribution of data to learn robust hash codes for retrieval systems. Although hierarchical semantic structures can be simply expressed by integrating semantically relevant data into a high-level taxon with coarser-grained semantics, the construction, embedding, and exploitation of the structures remain tricky for unsupervised hash learning. 
To tackle these problems, we propose a novel unsupervised hashing method named \textbf{H}yperbolic \textbf{H}ierarchical \textbf{C}ontrastive \textbf{H}ashing (\ourmethod). We propose to embed continuous hash codes into hyperbolic space for accurate semantic expression since embedding hierarchies in hyperbolic space generates less distortion than in hyper-sphere space and Euclidean space. In addition, we extend the K-Means algorithm to hyperbolic space and perform the proposed hierarchical hyperbolic K-Means algorithm to construct hierarchical semantic structures adaptively. To exploit the hierarchical semantic structures in hyperbolic space, we designed the hierarchical contrastive learning algorithm, including hierarchical instance-wise and hierarchical prototype-wise contrastive learning. Extensive experiments on four benchmark datasets demonstrate that the proposed method outperforms the state-of-the-art unsupervised hashing methods. Codes will be released.
\end{abstract}



\maketitle
\begin{figure}[t]
    \centering
    \includegraphics[width=\linewidth]{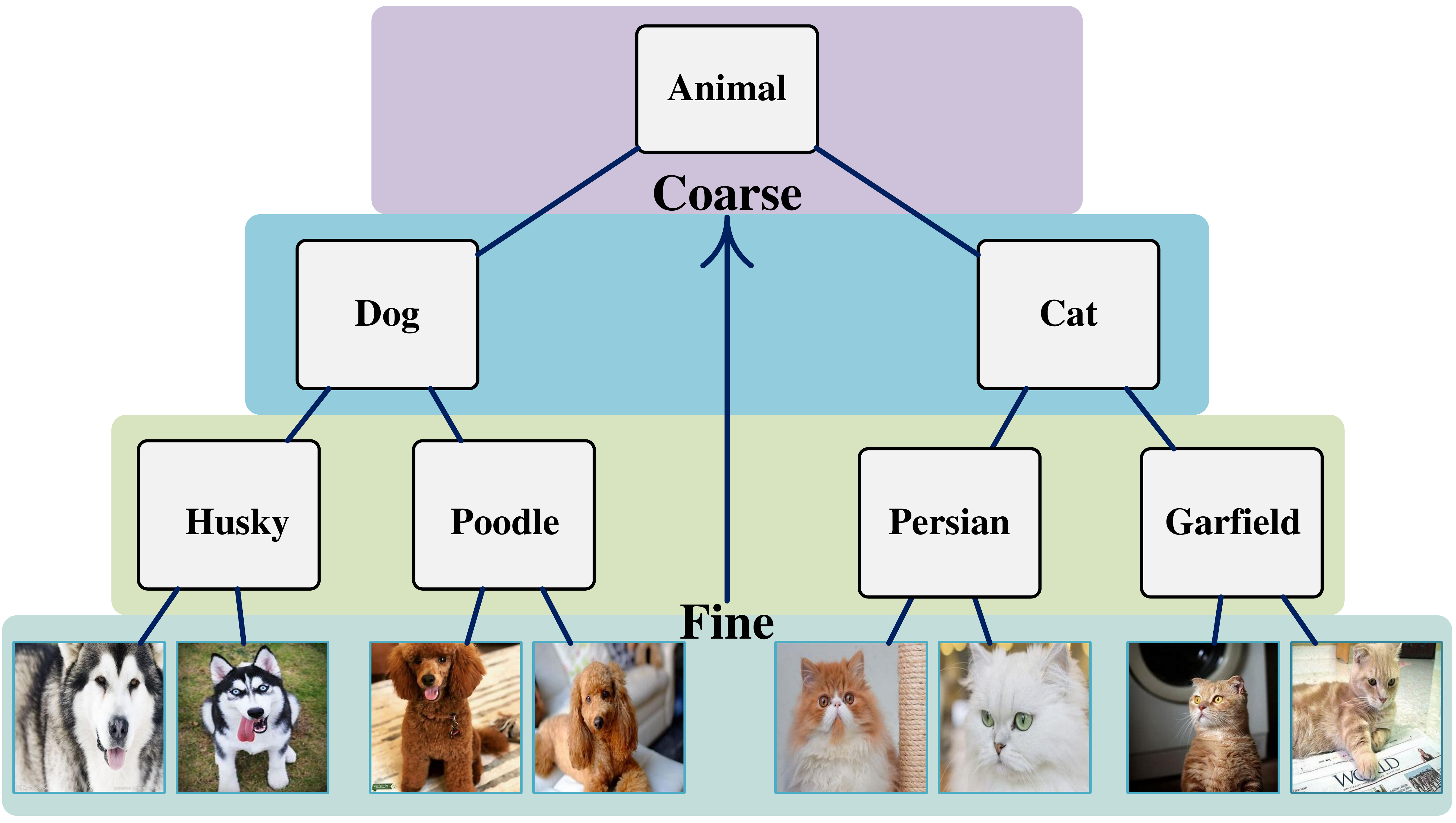}
    \caption{An illustration of a hierarchical semantic structure. Semantic hierarchy is an inherent property of real-world image datasets, \eg, "image instance $\rightarrow$ husky $\rightarrow$ dog $\rightarrow$ animal" in the order from fine-grained to coarse-grained semantics.}
    \label{fig:motivation_fig}
  \vspace{-0.4cm}
\end{figure}

\section{Introduction}
The explosive growth of multimedia data poses a huge challenge to large-scale information retrieval systems. Hashing-based methods~\cite{DCH2018CVPR,MLS3RDUH2020IJCAI,CSQ2020CVPR,BGAN2018AAAI,SSDH2018IJCAI,TBH2020CVPR,Hash_survey2018TPAMI}, converting high-dimensional features to compact binary hash codes while preserving the original similarity information in Hamming space, have become the dominant solution due to their high computation efficiency and low storage cost. Recently, unsupervised hashing methods~\cite{MISH2021WWW,BGAN2018AAAI,SSDH2018IJCAI,DistillHash2019CVPR,TBH2020CVPR,CIBHash2021IJCAI,MLS3RDUH2020IJCAI,BitEntropy2021AAAI} have attracted increasing attention since they do not rely on expensive hand-crafted labels and can perceive the distribution of target datasets for real-world retrieval tasks.

Contrastive hashing~\cite{DATE2021ACMMM,CIBHash2021IJCAI}, as the state-of-the-art unsupervised hashing method, learns hash codes by maximizing the mutual information between different views augmented by an image. However, none of them explores how hierarchical semantics can be used to improve the quality of hash codes, even though the hierarchical semantic structure as an inherent property of image datasets can assist in capturing the latent data distribution. As shown in Figure~\ref{fig:motivation_fig}, $Animal$ is a high-level taxon compared to $Dog$. We can obtain a tree-like hierarchical structure with an increasingly coarse semantic granularity from bottom to top.
Recently, Lin~\etal~\cite{DSCH2022AAAI} use homology relationships over a two-layer semantic structure to learn hash codes. Although employing the two-layer structure achieves excellent results, there is still room for improvement in constructing and exploiting the hierarchical semantic structures, acquiring more accurate cross-layer affiliation and cross-sample similarity. In addition, embedding the hierarchies into Euclidean or hyper-sphere spaces may miss the optimal solution due to information distortion~\cite{Hyper2020CVPR,HyperViT2022CVPR,HyperSurvey2021}. 

To address these problems, we propose constructing hierarchical structures and learning hash codes in hyperbolic space (\eg, the Poincar\'e ball). Since hyperbolic space has exponential volume growth with respect to the radius~\cite{HyperViT2022CVPR,Hyper2020CVPR,HyperSurvey2021} and can use low-dimensional manifolds for embeddings without sacrificing the model's representation power~\cite{hyper_acc2017NIPS,HyperViT2022CVPR}, it results in a lower distortion for embedding hierarchical semantics than Euclidean space with polynomial growth~\cite{hyper_low2011GD}. To achieve the construction of hierarchical semantic structures in hyperbolic space, we designed the \emph{hierarchical hyperbolic K-Means} algorithm. The algorithm performs bottom-up clustering with instances over the bottom layer and with prototypes over other layers by the hyperbolic K-Means algorithm, where the hyperbolic K-Means algorithm is our extended K-Means algorithm~\cite{kmeans1967SomeMF} from Euclidean space to hyperbolic space (See $\S$~\ref{sec:HH_K-Means}). In addition, referring to~\cite{HCSC2022CVPR}, we propose a hierarchical contrastive learning framework for hashing, including hierarchical instance-wise contrastive learning and hierarchical prototype-wise contrastive learning (See $\S$~\ref{sec:HCL}). The former leverages hierarchical semantic structures to mine accurate cross-sample similarity, reducing the number of \emph{false negatives}~\cite{negative2022AAAI,Negative2022ACL,negative2022ICML} to improve the discriminating ability. The latter aligns the hash codes of image instances with the corresponding prototypes (hash centers~\cite{CSQ2020CVPR}) over different layers, mining accurate cross-layer affiliation.

Based on these improvements, we propose a novel unsupervised hashing method called \textbf{H}yperbolic \textbf{H}ierarchical \textbf{C}ontrastive \textbf{H}ashing (\ourmethod). In the \ourmethod framework, we learn continuous hash codes and embed them into hyperbolic space with a projection head (See $\S$~\ref{sec:HSL}). Meanwhile, we perform hierarchical hyperbolic K-Means over the hyperbolic embeddings to construct hierarchical semantic structures before each training epoch. For hash learning within a mini-batch, we employ the proposed hierarchical contrastive learning under the SimCLR~\cite{SimCLR2020ICML} framework with the captured hierarchies. Finally, we conducted extensive experiments on four benchmark datasets to verify the superiority of \ourmethod compared with several state-of-the-art unsupervised hashing methods. The experimental results demonstrate that learning hash codes with the construction, embedding, and exploitation of hierarchical semantic structures in hyperbolic space can significantly improve retrieval performance.

Our main contributions can be outlined as follows:
\begin{itemize}
    \item We propose a novel contrastive hashing method named \ourmethod using the proposed hierarchical contrastive learning framework. The framework can benefit from the hierarchical semantic structures to improve the accuracy of cross-sample similarity and cross-layer affiliation for hash learning.
    \item We propose to project continuous hash codes into hyperbolic space (\ie, the Poincar\'e ball) for low information distortion. To this end, we designed the hierarchical hyperbolic K-Means algorithm that can work in hyperbolic space and adaptively construct hierarchical semantic structures from bottom to top.
    \item Extensive experiments on four benchmark datasets demonstrate that \ourmethod achieves superior retrieval performance compared with several state-of-the-art unsupervised hashing methods.
\end{itemize}

\section{Related Work}
\textbf{Unsupervised Hashing.} Existing unsupervised hashing methods mainly fall into two lines: reconstruction-based hashing methods and contrastive hashing methods. The former~\cite{SGH2017ICML,TBH2020CVPR,DVB2019IJCV} mostly adopts an encoder-decoder architecture~\cite{GAN2014NIPS,autoEncoding2013auto} to reconstruct original images from hash codes and others employ generative adversarial networks to maximize reconstruction likelihood via the discriminator~\cite{BGAN2018AAAI,BinGAN2018NIPS,HashGAN2018CVPR}. The latter can learn distortion-invariant hash codes, alleviating the problem of background noise caused by the reconstruction process and yielding state-of-the-art performance. Specifically, DATE~\cite{DATE2021ACMMM} proposes a general distribution-based metric to depict the pairwise distance between images, exploring both semantic-preserving learning and contrastive learning to obtain high-quality hash codes. CIBHash~\cite{CIBHash2021IJCAI} learns hash codes under the SimCLR~\cite{SimCLR2020ICML} framework and compresses the model by the Information Bottleneck~\cite{IB2015ITW}. Despite their contributions to learning compact hash codes in an unsupervised manner, they overlook rich information from hierarchical semantic structures inherent to the datasets. DSCH~\cite{DSCH2022AAAI} is aware of hierarchical semantics and tries to exploit them using homology and co-occurrence relationships mined by its two-step iterative algorithm. There is still room for the exploration of hierarchical semantics. 1) The customized two-layer hierarchical structure can only represent limited hierarchical information. It lacks an effective learning mechanism to adaptively construct hierarchical structures and provide accurate cross-sample and cross-layer information. 2) Embedding the hierarchical semantic structures in hyper-sphere space is not the optimal solution due to information distortion~\cite{Hyper2020CVPR,HyperViT2022CVPR,HyperSurvey2021}.


\noindent\textbf{Hyperbolic Embedding.} Recently, hyperbolic embedding technology has been successfully applied to CV~\cite{Hyper2020CVPR,HyperViT2022CVPR,HyperML2021CVPR} and NLP~\cite{NLP2017NPIS,NLP2018ACL,NLP2019ICLR} tasks due to the distinctive property of hyperbolic space, \ie, the exponential volume growth with respect to the radius rather than the polynomial growth in Euclidean space. Although it has been proven to be suitable for embedding hierarchies (\eg, tree graphs) with low distortion~\cite{HyperViT2022CVPR,HyperSurvey2021}, the algorithm for the construction of hierarchies in hyperbolic space has not been studied. We still need to explore construction schemes for hierarchical information in hyperbolic space for hashing tasks. For more details about hyperbolic embedding, we refer readers to~\cite{HyperSurvey2021} for a recent survey.

\noindent\textbf{Contrastive Learning.} Contrastive learning learns view-invariant representations by attracting positive samples and repelling negative samples. The \emph{instance-wise contrastive learning} methods~\cite{SimCLR2020ICML, MoCo2020CVPR, BYOL2020NIPS, SimSiam2021CVPR}, \eg, SimCLR~\cite{SimCLR2020ICML} and MoCo~\cite{MoCo2020CVPR}, maximize the identical representation between views augmented from the same instance. They highlight data-data correlations and neglect the global distribution of the whole dataset. To compensate for this, the \emph{prototype-wise contrastive learning} methods~\cite{PCL2021ICLR,SwAV2020NIPS,LLP-PCC2022AAAI,CC2021AAAI} explicitly exploit the semantic structure and learn the prototypes (\ie, the centers) of each cluster formed by semantically similar instances. Our \ourmethod benefits from both two kinds of contrastive learning methods as well as recent efforts in hierarchical representation learning~\cite{HCSC2022CVPR,GraphHC2021ICML}.

\begin{figure*}[ht]
    \centering
    \includegraphics[width=\textwidth,height=0.3\textwidth]{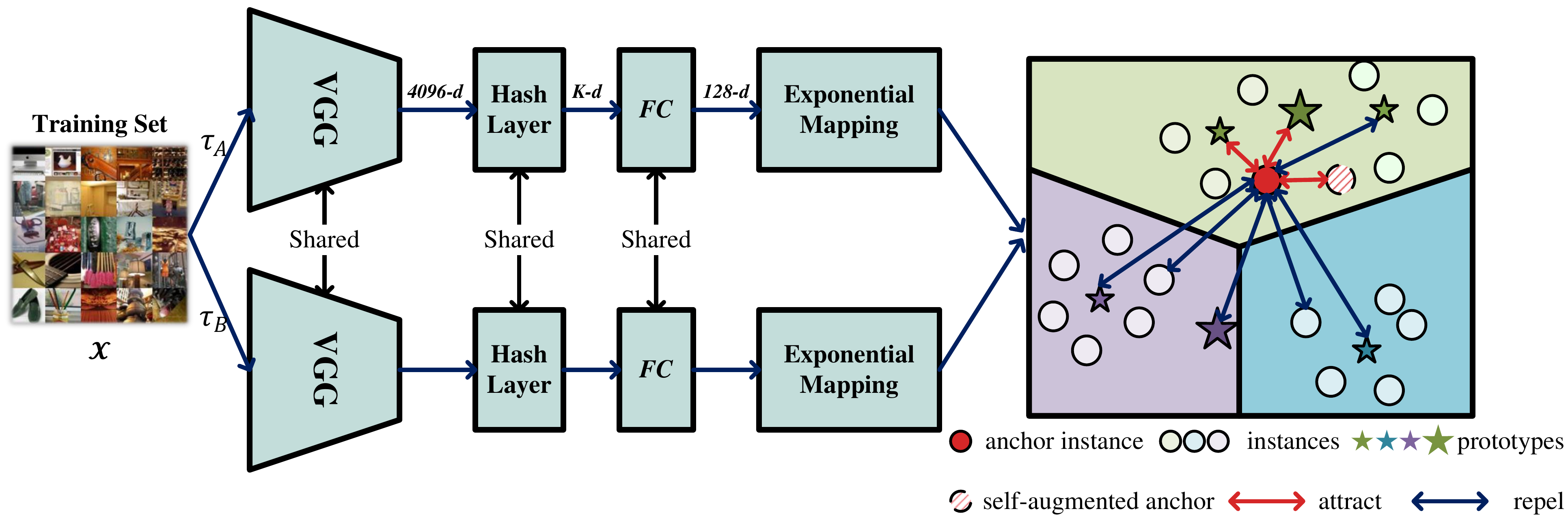}
    \caption{The framework of \ourmethod. $\tau_{A}$ and $\tau_{B}$ are augmentations used to transform an image. Each transformed view is encoded by VGG and transformed into a $K$-bit continuous hash code. Then, the code is projected into hyperbolic space by a projection head consisting of a fully connected layer and an exponential mapping function. \ourmethod captures the hierarchical semantic structures before each training epoch and conducts hierarchical contrastive learning to exploit hierarchical information in hyperbolic space. In hierarchical contrastive learning, the anchor sample will contrast with both instances and prototypes at different layers.}
    \label{fig:framework}
    \vspace{-1.8em}
\end{figure*}

\section{Methodology}

\subsection{Problem Definition and Overview}
Given a training set $\mathcal{X}=\left\{x_i\right\}_{i=1}^{N}$ of $N$ unlabeled images, we aim to learn a nonlinear hash function $f_{\theta_{h}}: x\rightarrow h\in\left\{-1,1\right\}^{K}$ that maps the data from input space $\mathbb{R}^D$ to $K$-bit Hamming space. The learning procedure and details of \ourmethod are shown in Algorithm~\ref{alg:algorithm_HHCH}. 

In the training phase, given an image $x_{i}$, our proposed $f_{\theta_{h}}$ extracts the feature vector using the VGG model~\cite{vgg2014ICLR} and generates the continuous hash code $h_{i}$ through the hash layer. Then, $h_{i}$ is projected to $z_{i}$, \ie, embedding the hash code into hyperbolic space (\ie, the Poincar\'e ball), with a projection head $Exp\_map_{\theta_{e}}$. The projection head contains a fully-connected layer followed by the exponential mapping function in Equation (\ref{eq:exp}).
\begin{algorithm}[t]
	\caption{HHCH algorithm.}
	\label{alg:algorithm_HHCH}
	\KwIn{Training data $\mathcal{X}$; batch size $B$; Hyper-parameters; Training epochs $T$; Hash function $f_{\theta_{h}}$ and projection head $Exp\_map_{\theta_{e}}$.}
	\KwOut{Optimized hash function $f_{\theta_{h}}$}  
	\BlankLine
	Initialize $\bm{\theta_{h}}$ and $\bm{\theta_{e}}$ randomly;

	\For{$t$=1 to $T$}{
	    $Z=Exp\_map_{\theta_{e}}(f_{\theta_{h}}(\mathcal{X}))$;
	    
	    \tcc{Construct the hierarchical semantic structures and output prototypes $\mathcal{P}$ and connections $\mathcal{E}$.}
		$(\mathcal{P},\mathcal{E})$ $\leftarrow$ Hierarchical hyperbolic  K-Means($Z$);
		
		\For{$b=1$ to $\frac{N}{B}$}{
		
		$X_{b}^{1}$, $X_{b}^{2} \leftarrow transformation(X_{b}) $
		
		$H_{b}^{1}, H_{b}^{2}$=$f_{\theta_{h}}(X_{b}^{1})$, $f_{\theta_{h}}(X_{b}^{2})$
		
		$Z_{b}^{1}, Z_{b}^{2}$=$Exp\_map_{\theta_{e}}(H_{b}^{1})$, $Exp\_map_{\theta_{e}}(H_{b}^{2})$;
		
		Compute $L_{H-inst}$ and $L_{H-proto}$ with  $(Z_{b}^{1},Z_{b}^{2},\mathcal{P},\mathcal{E})$;
		
		Compute $L_{Q}$ with  $(H_{b}^{1},H_{b}^{2})$;
		
		Compute $\mathcal{L}$ with Equation (16);
		
		Update $\bm{\theta_{h}}$ and $\bm{\theta_{e}}$ with the Adam optimizer.
		}
	}
\end{algorithm}
Before each training epoch, we generate the hyperbolic embeddings $Z=\left\{z_{i}\right\}_{i=1}^{N}$ of all training images and perform \emph{hierarchical hyperbolic K-Means} to construct hierarchical semantic structures. For the hash learning within the $b$-th mini-batch, we transform every image into two views with various augmentation strategies. Then, we acquire the corresponding hash codes $H_{b}^{1}=\left\{h_{i}^{1}\right\}_{i=1}^{B}$ and $H_{b}^{2}=\left\{h_{i}^{2}\right\}_{i=1}^{B}$ as well as the hyperbolic embeddings $Z_{b}^{1}=\left\{z_{i}^{1}\right\}_{i=1}^{B}$ and $Z_{b}^{2}=\left\{z_{i}^{1}\right\}_{i=1}^{B}$, where $B$ denotes the batch size. Finally, we compute the hierarchical contrastive loss, including the hierarchical instance-wise contrastive loss and the hierarchical prototype-wise contrastive loss, with the hyperbolic embeddings and the captured hierarchies $(\mathcal{P},\mathcal{E})$, where $\mathcal{P}$ is the set of prototypes and $\mathcal{E}$ is the set of connections consisting of prototype-prototype and prototype-instance. In addition, we incorporate the quantization loss to reduce the quantization error.

In the test phase, we only use the well-trained hash function $f_{\theta_{h}}$, disabling the projection head and the hierarchical semantic structures. All the continuous hash codes will be constrained to $\left\{-1,1\right\}^{K}$ by the $sgn$ function for performance evaluation, where $K$ denotes the length of the hash code.

The framework of \ourmethod is shown in Figure~\ref{fig:framework}. In the following subsections, we will specify \ourmethod by answering the questions below.

\noindent\textbf{Q1:} How can we bridge Euclidean space and hyperbolic space?

\noindent\textbf{Q2:} How to construct hierarchical semantic structures in hyperbolic space?

\noindent\textbf{Q3:} How to exploit hierarchical semantic structures for hash learning?

\subsection{Hyperbolic Space Learning (RQ1)}
\label{sec:HSL}
Formally, $n$-dimensional hyperbolic space $\mathbb{H}^n$ is a Riemannian manifold of constant negative curvature rather than the constant positive curvature in Euclidean space. There exist several isomorphic models of hyperbolic space, we specialize in the Poincar\'e ball model $(\mathbb{D}^n_c, g^{\mathbb{D}})$ with the curvature parameter $c$ (the actual curvature value is then $-c^2$) in this work. The model is defined by the manifold $\mathbb{D}^n = \{ x \in \mathbb{R}^n \colon c\|x\|^2 < 1, c \geq 0\} $ endowed with the Riemannian metric $g^{\mathbb{D}} = \lambda_c^2 g^E$, where $ \lambda_c = \frac{2}{1-c\|x\|^2}$ is the conformal factor and $g^E = \mathbf{I}_n$ is the Euclidean metric tensor~\cite{Hyper2020CVPR,HyperSurvey2021,HyperViT2022CVPR}. 

Since hyperbolic space is not vector space in a traditional sense, we must introduce the gyrovector formalism~\cite{hyper_gyrovector2008Ungar} to perform operations such as addition~\cite{Hyper2020CVPR}. As a result, we can define the following operations in Poincar\'e ball:

\noindent\textbf{M\"obius addition.} For a pair $\mathbf{x}, \mathbf{y} \in \mathbb{D}^n_c$, their addition is defined below.
\begin{equation}
    \mathbf{x} \oplus_c \mathbf{y} = \frac{(1+2c\langle \mathbf{x}, \mathbf{y} \rangle + c\|\mathbf{y}\|^2) \mathbf{x}+ (1-c\|\mathbf{x}\|^2)\mathbf{y}}{1+2c\langle \mathbf{x}, \mathbf{y} \rangle + c^2 \|\mathbf{x}\|^2 \|\mathbf{y}\|^2}.
\end{equation}

\noindent\textbf{Hyperbolic distance.} The hyperbolic distance between $\mathbf{x}, \mathbf{y} \in \mathbb{D}^n_c$ is defined below.
\begin{equation}
\label{eqn:distance_hyper}
\begin{aligned}
 D_{hyp}(\mathbf{x},\mathbf{y}) = \frac{2}{\sqrt{c}} \mathrm{arctanh}(\sqrt{c}\|-\mathbf{x} \oplus_c \mathbf{y}\|).
\end{aligned}
\end{equation}
Note that with $c \to 0$, the distance function \eqref{eqn:distance_hyper} reduces to the Euclidean distance: $\lim_{c \to 0} D_{hyp}(\mathbf{x},\mathbf{y})=2\|\mathbf{x}-\mathbf{y}\|.$ 

\noindent\textbf{Exponential mapping function.} We also need to define a bijective map from Euclidean space to the Poincar\'e model of hyperbolic geometry. This mapping is termed \textit{exponential},  while its inverse mapping from hyperbolic space to Euclidean is called \textit{logarithmic}. For some fixed base point $\mathbf{x} \in \mathbb{D}^n_c$, the exponential mapping is a function $\exp_\mathbf{x}^c \colon \mathbb{R}^n \to \mathbb{D}_c^n$ that is defined as follows:
\begin{equation}\label{eq:exp}
    \exp_\mathbf{x}^c(\mathbf{v}) = \mathbf{x} \oplus_ c \bigg(\tanh \bigg(\sqrt{c} \frac{\lambda_\mathbf{x}^c \|\mathbf{v}\|}{2} \bigg) \frac{\mathbf{v}}{\sqrt{c}\|\mathbf{v}\|}\bigg).
\end{equation}
Usually, the base point $\mathbf{x}$ is set to $\mathbf{0}$, making the above formulas simple but with little bias to the original results~\cite{HyperViT2022CVPR}.

In hyperbolic space, the local distances are scaled by the factor $\lambda_c$, approaching infinity near the boundary of the ball. As a result, hyperbolic space has the ``space expansion property''. While in Euclidean space, the volume of an object with a diameter of $r$ scales polynomially with $r$, in hyperbolic space, the counterpart scales \emph{exponentially} with $r$. Intuitively, this is a continuous analog of trees: for a tree with a branching factor $k$, we obtain $O(k^d)$ nodes on level $d$, which in this case serves as a discrete analog of the radius. This property allows us to efficiently embed hierarchical data even in low dimensions, which is made precise by embedding theorems for trees and complex networks~\cite{hyper_low2011GD,HyperViT2022CVPR}.

\subsection{Hierarchical Hyperbolic K-Means (RQ2)}
\label{sec:HH_K-Means}
We aim to construct hierarchical structures by capturing hierarchical relationships among semantic clusters in hyperbolic space.
To this end, we propose the \emph{hierarchical hyperbolic K-Means} algorithm in the Poincar\'e ball, constructing the structures in a bottom-up manner.

The details of \emph{hierarchical hyperbolic K-Means} are shown in Algorithm~\ref{alg:algorithm_HHK_Means}. We define the number of prototypes at the $l$-th layer as $M_{l}$ and the total number of layers as $L$. First, we obtain the hyperbolic embeddings $Z=\{z_{i}\}_{i=1}^{N}$ of all images before each training epoch. Then, we perform \emph{hyperbolic K-Means} with $Z$ to obtain the prototypes of the first/bottom layer, \ie, $\{p_{j}^{1}\}_{j=1}^{M_{1}}$. Similarly, the prototypes of each higher layer are derived by iteratively applying hyperbolic K-Means to the prototypes of the layer below~\cite{HCSC2022CVPR}. We record the hierarchical information by maintaining the prototype set $\mathcal{P}=\{\{p_{j}^{l}\}_{j=1}^{M_{l}}\}_{l=1}^{L}$ and the connection set $\mathcal{E}$. 

In the above process, the \emph{hyperbolic K-Means} algorithm is the key to achieving the construction in hyperbolic space. Although the K-Means algorithm optimizes the prototypes and the latent cluster assignments alternatively, existing variants of K-Means define the prototype by the Euclidean averaging operation over all embeddings within the cluster, which does not apply to hyperbolic space. To perform K-Means in hyperbolic space, we have to define 1) the distance metric in hyperbolic space and 2) the calculation of prototypes. The former has been solved in Equation (\ref{eqn:distance_hyper}). For the latter, we compute the \textbf{Einstein midpoint} as the prototype in hyperbolic space, referring to~\cite{HAN2019ICLR}, which has the most concise form with the Klein coordinates~\cite{HyperSurvey2021,Hyper2020CVPR} as follows:  
\begin{algorithm}[t]
	\caption{Hierarchical Hyperbolic K-Means.}
	\label{alg:algorithm_HHK_Means}
	\KwIn{Image hyperbolic embedding $Z$; Number of hierarchies $L$, Number of clusters at the $l$-th hierarchy $M_l$.}
	\KwOut{Hierarchical semantic structures $\left(\mathcal{P},\mathcal{E}\right)$.}  
	$\{p_{j}^{1}\}_{j=1}^{M_{1}} \leftarrow $ Hyperbolic K-Means($Z$)
	
	$\mathcal{E}=\{\left(z_{i}, Parent(z_{i})\right)\}_{i=1}^{N}$
	
	\For{$l$=2 to $L$}{
		$\{p_{j}^{l}\}_{j=1}^{M_{l}} \leftarrow$ Hyperbolic K-Means($\{p_{j}^{l-1}\}_{j=1}^{M_{l-1}}$)
		
		\tcc{Update connections. $Parent(\cdot)$ means the parent node at a higher level.}
		
		$\mathcal{E} \leftarrow \mathcal{E}\cup \{(p_{j}^{l-1},Parent(p_{i}^{l-1}))\}_{j=1}^{M_{l-1}}$
		
	}
	\BlankLine
	\BlankLine
    \tcc{ More details about Hyperbolic K-Means} 
    
    Initial prototypes using K-Means++ in hyperbolic space with the distance calculated by Equation (2).
    
    \While{\textnormal{not converged}}{
    
    calculate cluster assignments according to Equation (2);
    
    \tcc{The following is the calculation of new prototypes.} 
    
    Map all points to the Klein model with Equation (6);
    
    Compute Einstein midpoints with Equation (4);
    
    Project all midpoints (prototypes) back to the Poincar\'e  ball;
    
    }

\end{algorithm}

\begin{equation}
\label{eqn:cal_proto}
    \begin{aligned}
        Klein\_Proto(x_1,...,x_{N})=\sum_{i=1}^{N}\gamma_{i}x_{i}/\sum_{i=1}^{N}\gamma_{i},
    \end{aligned}
\end{equation}
where $\gamma_{i}=\frac{1}{\sqrt{1-c||x_i||^2}}$ is the Lorentz factor~\cite{HyperSurvey2021}. Since the Klein model and the Poincar\'e ball model are isomorphic~\cite{Hyper2020CVPR}, we can transition between $x_{\mathbb{D}}$ in the Poincar\'e ball and $x_{\mathbb{K}}$ in the Klein model as follows:
\begin{equation}
    \begin{aligned}
    x_{\mathbb{K}}=\frac{2x_{\mathbb{D}}}{1+c||x_{\mathbb{D}}||^2},
    \end{aligned}
\end{equation}

\begin{equation}
    \begin{aligned}
    x_{\mathbb{D}}=\frac{x_{\mathbb{K}}}{1+\sqrt{1-c||x_{\mathbb{K}}||^2}}.
    \end{aligned}
\end{equation}

Based on these formulas, we can map all the points in the Poincar\'e ball to the Klein model, computing the prototypes via Equation (\ref{eqn:cal_proto}) first and then projecting them back to the Poincar\'e model. As a result, we can conduct hyperbolic K-Means clustering via alternative optimization with the distance metric and the prototype calculation like the existing K-Means algorithms.
\subsection{Hierarchical Contrastive Learning (RQ3)}
\label{sec:HCL}
Since the contrastive hashing methods~\cite{CIBHash2021IJCAI,DSCH2022AAAI,DATE2021ACMMM} have achieved satisfying retrieval performance without hand-crafted labels, we adopt the classic contrastive learning framework SimCLR~\cite{SimCLR2020ICML} as our fundamental learning framework. To incorporate hierarchical information into the contrastive hashing framework, we propose hierarchical instance-wise contrastive learning and hierarchical prototype-wise contrastive learning, as well as extending them to work in hyperbolic space. We will elaborate on how contrastive hashing benefits from hierarchical semantic structures using both learning patterns.

\noindent\textbf{Hierarchical instance-wise contrastive learning.}
Instance-wise contrastive learning pushes the embeddings of two transformed views of the same image (positives) close to each other and further apart from the embeddings of other images (negatives), where the instance-wise contrastive loss is defined as follows:

\begin{equation}
\label{eqn:contrastive_instanct}
    \small
   \begin{aligned}
    \Tilde{\ell}_{i}^{1}=-\log\frac{exp\left(-D\left(z_{i}^{1},z_{i}^{2}\right)/\tau\right)}{exp\left(-D\left(z_{i}^{1},z_{i}^{2}\right)/\tau \right)+\sum^{}_{v,z_n\in \mathcal{N}(z_i)} exp\left(-D\left(z_{i}^{1},z_{n}^{v}\right)/\tau\right)},
    \end{aligned}
\end{equation}
where $z_{i}^{v}$ is the representation of the $v$-th view of $x_{i}$, $\Tilde{\ell}_{i}^{v}$ denotes the instance-wise contrastive loss of $z_{i}^{v}$, $\mathcal{N}(z_{i})$ is the negative set of $z_{i}$, $\tau$ is the temperature parameter~\cite{SimCLR2020ICML,MoCo2020CVPR}, and $D$ is the distance metric. We can express $D$ by the cosine distance defined in hyper-sphere space, \ie,
\begin{equation}
\label{eqn:cos_dist}
    \begin{aligned}
    D_{cos}\left(z_{i},z_{j}\right)=\frac{K}{2}\left(1-cos\left(z_{i},z_{j}\right)\right).
    \end{aligned}
\end{equation}
Similarly, $D$ can be defined as the hyperbolic distance in Equation (\ref{eqn:distance_hyper}). As a result, the instance-wise contrastive loss for all views of instance $z_{i}$ is formulated as
\begin{equation}
\label{eqn:cl_all_view}
    \begin{aligned}
    L_{inst}\left(z_{i},\mathcal{N}(z_{i})\right)=\sum_{v=1}^{2}\Tilde{\ell}_{i}^{v}.
\end{aligned}
\end{equation}

Recent studies show that the selection of negative samples is critical for the quality of contrastive learning~\cite{SimCLR2020ICML,negative2022AAAI,Negative2022ACL,negative2022ICML}. Existing contrastive hashing methods~\cite{CIBHash2021IJCAI,DATE2021ACMMM,DSCH2022AAAI} suffer from the false negative problem~\cite{Negative2022ACL,negative2022ICML} since they treat all the remaining images $z_{j}^{v}$ within a mini-batch as negatives when given a random anchor image $z_{i}^{v}$, even if the negative images share the same semantic as the anchor image. We aim to sample distinctive negative samples according to the hierarchical semantic structure, alleviating the problem to achieve a solid discriminating ability for the contrastive hashing model. 



Compared to flat structures, hierarchical structures can provide hierarchical similarity based on affiliation, resulting in accurate cross-sample similarity for negative sampling. Specifically, we sample negative samples according to the hierarchical structure constructed in $\S$~\ref{sec:HH_K-Means}. Given the anchor sample $z_{i}$, the negative sample set $\mathcal{N}^{l}(z_{i})$ at the $l$-th layer can be defined by

\begin{equation}
    \small
    \begin{aligned}
    \mathcal{N}^{l}\left(z_{i}\right)=\left\{z_{j} \mid j=1,..., B \text{ and } \mathcal{P}^{l}(z_{j}) \!\neq\! \mathcal{P}^{l}(z_{i}) \right\},
    \end{aligned}
\end{equation}
where $\mathcal{P}^{l}(z_{i})$ is the ancestor of $z_{i}$ at the $l$-th layer. The above equation implies that the negative sample set of the anchor sample $z_{i}$ only contains samples with a different ancestor than $z_{i}$ at the $l$-th layer.


The negative sample sets at different levels in hierarchical contrastive hashing take on varying importance because the granularity of the hierarchical semantic structure decreases from fine to coarse as the level advances. As a result, the overall hierarchical instance-wise contrastive learning objective $\mathcal{L}_{H-inst}$ can be formulated as a weighted summation of contrastive loss in Equation (\ref{eqn:cl_all_view}) at different layers, \ie,

\begin{equation}
    \begin{aligned}
    \mathcal{L}_{H-inst}=\frac{1}{B\cdot L}\sum_{i=1}^{B}\sum_{l=1}^{L}\frac{1}{l}L_{inst}\left(z_{i}, \mathcal{N}^{l}(z_{i})\right).
    \end{aligned}
\end{equation}

\noindent\textbf{Hierarchical prototype-wise contrastive learning.}
Compared to instance-wise contrastive learning methods~\cite{SimCLR2020ICML,MoCo2020CVPR,BYOL2020NIPS} that learn data-data correlations, contrasting instance-prototype pairs can capture the global data distribution to acquire accurate cross-layer affiliation.
To this end, we introduce prototype-wise contrastive learning~\cite{PCL2021ICLR}, where we define the ancestor $\mathcal{P}^{l}(z_{i})$ of $z_{i}$ as the positive sample and all the remaining prototypes as negative samples. Analogous to Equation (\ref{eqn:contrastive_instanct}), the prototype-wise contrastive loss is defined below.

\begin{equation}
\label{eqn:contrastive_proto}
    \small
   \begin{aligned}
    \hat{\ell}_{i}^{v}\!\!=-\!\log\!\frac{exp\left(-\!D\left(z_{i}^{v},\mathcal{P}(z_{i})\right)\!/\tau\right)}{exp\left(-\!D\left(z_{i}^{v},\mathcal{P}(z_{i})\right)\!/\tau \right)\!+\!\!\sum_{p_n\in \mathcal{N}_{p}(z_i)}\! exp\left(-\!D(z_{i}^{v},p_{n})\!/\tau\right)},
\end{aligned}
\end{equation}
where $N_p(z_{i})$ is the negative prototypes set of $z_{i}$ and $p_n \in N_p(z_{i})$. The prototype-wise contrastive loss for all views can be defined as follows:
\begin{equation}
    \begin{aligned}
     L_{proto}(z_{i},\mathcal{N}_{p}(z_{i}))=\sum_{v=1}^{2}\hat{\ell}_{i}^{v}.
    \end{aligned}
\end{equation}
We employ a different negative sampling strategy for the hierarchical prototype-wise contrastive learning than for the hierarchical instance-wise learning for two reasons. On the one hand, since negative prototypes are usually distinct from the anchor image, prototype-wise contrastive learning suffers less from the false negative problem than  instance-wise contrastive learning. On the other hand, since the number of prototypes is much less than the number of instances, the negative sampling strategy for hierarchical instance-wise contrastive learning will cause the problem of under-sampling for negatives in hierarchical prototype-wise contrastive learning.

By contrasting prototypes at different layers, we define the hierarchical prototype-wise contrastive loss as follows:
\begin{equation}
    \begin{aligned}
     \mathcal{L}_{H-proto}=\frac{1}{B\cdot L}\sum_{i=1}^{B}\sum_{l=1}^{L}\frac{1}{l}L_{proto}\left(z_{i}, \mathcal{N}_{p}^{l}(z_{i})\right).
    \end{aligned}
\end{equation}

In addition, referring to~\cite{DHN2016AAAI}, we define the quantization loss to reduce the accumulated quantization error caused by the continuous relaxation, where the loss is defined below.
\begin{equation}
    \begin{aligned}
     \mathcal{L}_{Q}=\frac{1}{2}\sum_{i=1}^{B}\sum_{k=1}^{K}\sum_{v=1}^{2}\left(\log cosh(|h_{i,k}^{v}|-\mathbf{1}) \right),
    \end{aligned}
\end{equation}
where $h_{i,k}$ is the $k$-th bit of $h_{i}$ and $\mathbf{1} \in \mathbb{R}^{K}$ is the vector of ones. Finally, the overall learning objective of hierarchical contrastive hashing can be formulated as follows:
\begin{equation}
    \begin{aligned}
     \mathcal{L}=\mathcal{L}_{H-inst}+\mathcal{L}_{H-proto}+\lambda\mathcal{L}_{Q}.
    \end{aligned}
\end{equation}
where $\lambda$ is the hyper-parameter to trade off different loss items.
\begin{table*}[t]
		\centering
		\caption{Comparison in mAP of Hamming Ranking for different bits on image retrieval.}
		
		\scalebox{0.9}{
			\begin{tabular}{llcccccccccccc}
				\toprule
				\multirow{2}{*}{Method} &
				\multirow{2}{*}{Reference} &
				\multicolumn{3}{c}{ImageNet}&
				 \multicolumn{3}{c}{CIFAR-10}&
				\multicolumn{3}{c}{FLICKR25K} &
				\multicolumn{3}{c}{NUS-WIDE} \\
				\cmidrule(lr){3-5}  \cmidrule(lr){6-8} \cmidrule(lr){9-11} \cmidrule(lr){12-14}
				& & 16-bit & 32-bit & 64-bit& 16-bit & 32-bit & 64-bit  & 16-bit & 32-bit  & 64-bit  & 16-bit & 32-bit  & 64-bit \\
				\midrule[0.8pt]
			    SGH~\cite{SGH2017ICML}&ICML17&0.557 &0.572 &0.583 & 0.286& 0.320& 0.347& 0.608& 0.657& 0.693& 0.463 &0.588 &0.638 \\
				SSDH~\cite{SSDH2018IJCAI}&IJCAI18&0.604&0.619 &0.631 &0.241& 0.239& 0.256& 0.710& 0.696& 0.737 &0.542& 0.629& 0.635  \\
				BGAN~\cite{BGAN2018AAAI}&AAAI18&0.649&0.665 &0.675   &  0.535& 0.575& 0.587& 0.766& 0.770& 0.795& 0.719& 0.745& 0.761 \\
				DistillHash~\cite{DistillHash2019CVPR}&CVPR19&0.654&0.671 &0.683 &0.547   &0.582  &0.591  &0.779  &0.793  &0.801  &0.722  &0.749  &0.762  \\
				ML$S^3$RDUH~\cite{MLS3RDUH2020IJCAI}&IJCAI20&0.662 &0.680 &0.691  & 0.562& 0.588& 0.595& 0.797& 0.809& 0.809& 0.730& 0.754 &0.764 \\ 
			    TBH~\cite{TBH2020CVPR}&CVPR20&0.636 &0.653 &0.667 & 0.432 & 0.459 & 0.455& 0.779& 0.794& 0.797& 0.678& 0.717& 0.729  \\ 
				CIBHash~\cite{CIBHash2021IJCAI}&IJCAI21 &0.719 &0.733 &0.747 &  0.547& 0.583& 0.602& 0.773 &0.781& 0.798& 0.756& 0.777& 0.781 \\ 
				DSCH~\cite{DSCH2022AAAI}&AAAI22&0.749 &0.761 &0.774 &   0.624 & 0.644 &0.670& 0.817& 0.827& 0.828& 0.770& 0.792& 0.801 \\ 
				\midrule[0.8pt]
				\bf \ourmethod  & \bf Ours &\textbf{ 0.783} & \textbf{0.814} & \textbf{0.826}   &\textbf{0.631}  &\textbf{0.657}  &\textbf{0.681}  &\textbf{0.825}  &\textbf{0.838}  &\textbf{ 0.842} &\textbf{0.797}  &\textbf{0.820}  &\textbf{0.828}   \\
				\bottomrule
			\end{tabular}
		}
		\label{table:all_map}
	\end{table*}
	

\section{Experiments}
In this section, we conduct experiments on four public benchmark datasets to evaluate the superiority of our proposed \ourmethod. More detailed experimental results and additional visualizations can be found in the \textbf{supplementary material}. Note that baseline results are reported from DSCH~\cite{DSCH2022AAAI}.
\subsection{Dataset and Evaluation Metrics}
The public benchmark datasets include ImageNet~\cite{ImageNet2009CVPR}, CIFAR-10~\cite{cifar2009learning}, FLICKR25K~\cite{flickr2008ICMR}, and NUS-WIDE~\cite{nuswide2009CIVR}.

\noindent\textbf{ImageNet} is a commonly used single-label image dataset. Following~\cite{CSQ2020CVPR,HashNet2017CVPR,DCH2018CVPR}, we randomly select 100 categories for the experiments. Besides, we use 5,000 images as the query set and the remaining images as the retrieval set, where we randomly select 100 images per category as the training set.

\noindent\textbf{CIFAR-10} consists of 60,000 images containing 10 classes. We follow the common setting~\cite{DSCH2022AAAI} and select 1,000 images (100 per class) as the query set. The remaining images are used as the retrieval set, where we randomly selected 1,000 images per class to form the training set.

\noindent\textbf{NUS-WIDE} is a multi-label dataset that contains 269,648 images from 81 classes. Following the commonly used setting~\cite{DSCH2022AAAI,CIBHash2021IJCAI,TBH2020CVPR}, we only use images selected from 21 most frequent classes. Besides, we sample 100 images per class as the query set and use the remaining as the retrieval set, where we randomly select 10,500 images (5,00 images per class) to form the training set.

\noindent\textbf{FLICKR25K} is a multi-label dataset containing 25,000 images from 24 categories. Following~\cite{DSCH2022AAAI}, we randomly sample 1,000 images as the query set, and the remaining images are left for the retrieval set. In the retrieval set, we randomly choose 10,000 images as the training set.

\noindent\textbf{Evaluation Protocol.} To evaluate retrieval quality, we follow~\cite{TBH2020CVPR,CSQ2020CVPR,DVB2019IJCV,MLS3RDUH2020IJCAI,PSLDH2021WWW} to employ the following metrics: 1) Mean Average Precision (mAP), 2) Precision-Recall (P-R) curves, 3) Precision curves w.r.t. different numbers of returned samples (P@N),  4) Precision curves within Hamming radius 2 (P@H$\le$2), 5) Mean intra-class distance $d_{intra}$, and 6) Mean inter-class distance $d_{inter}$. According to~\cite{CIBHash2021IJCAI,TBH2020CVPR}, we adopt mAP@1000 for ImageNet and CIFAR-10, as well as mAP@5000 for FLICKR25K and NUS-WIDE.
\begin{table}[t]
    \centering
    \caption{The $d_{intra}$ and $d_{inter}$ on ImageNet.}
    \scalebox{0.95}{
    \begin{tabular}{lcccccc}
    \toprule
         \multirow{2}{*}{Method}& 
         \multicolumn{3}{c}{$d_{intra} \downarrow$}&
         \multicolumn{3}{c}{$d_{inter} \uparrow$}\\
         \cmidrule(lr){2-4} \cmidrule(lr){5-7}
         &16-bit&32-bit&64-bit&16-bit&32-bit&64-bit\\ 
         \midrule
         CIBHash &1.23 &2.54 &5.12 & 8.04&16.08 &32.16 \\
         DSCH &1.03 &2.24 &4.59 &8.06  &16.12 &32.24\\
         \ourmethod(Ours) & \bf 0.81  & \bf1.88 & \bf 3.88& \bf 8.08 & \bf 16.15  &\bf 32.31\\
    \bottomrule
    \end{tabular}}
    
    \label{tab:distance_comparison}
\end{table}

\begin{figure}[ht]
    \centering
    \includegraphics[width=0.95\linewidth]{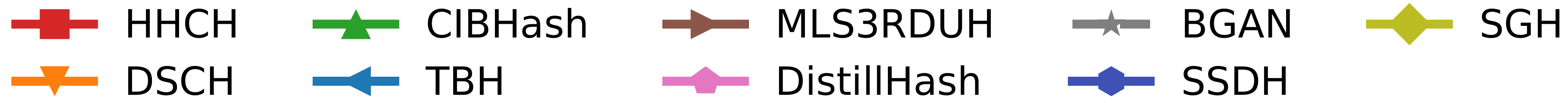}\\
    \subfloat[\footnotesize P-R curve at ImageNet:64bits]{\includegraphics[width=0.49\linewidth]{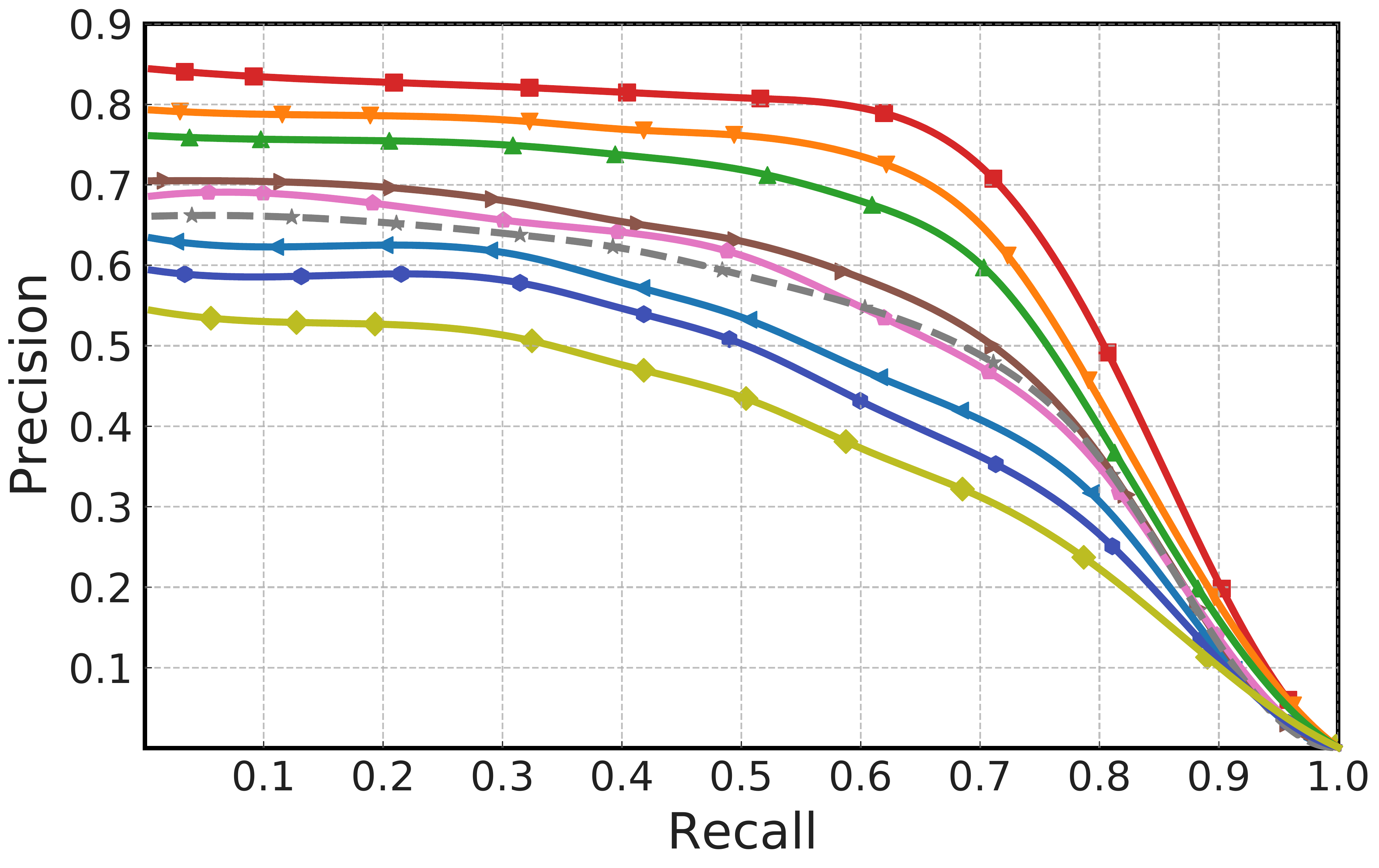}}
    \hfill
    \subfloat[\footnotesize  P-R curve at NUS-WIDE:64bits]{\includegraphics[width=0.49\linewidth]{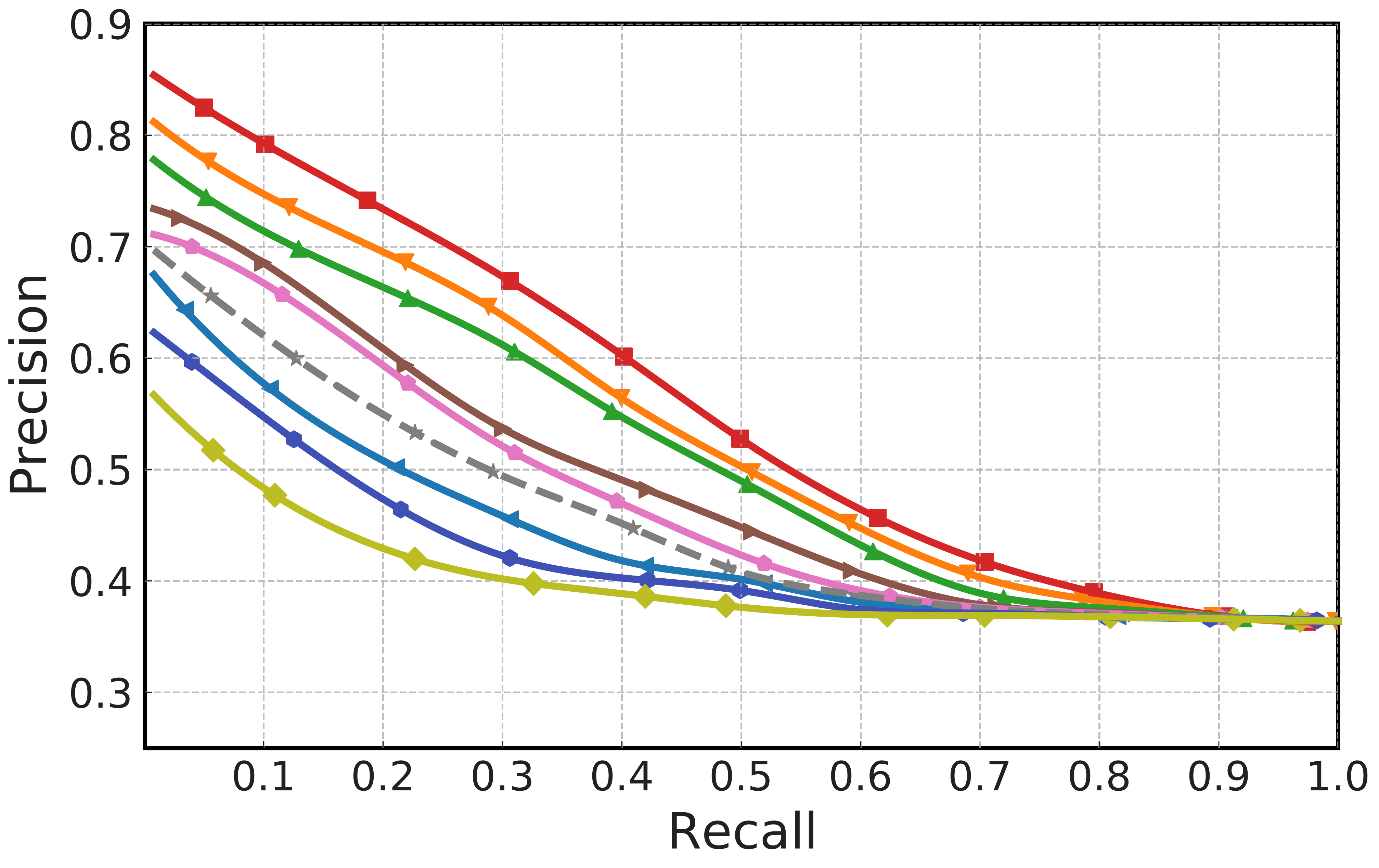}}
    \\
    \subfloat[\footnotesize  P@N=5000 at ImageNet:64bits]{\includegraphics[width=0.49\linewidth]{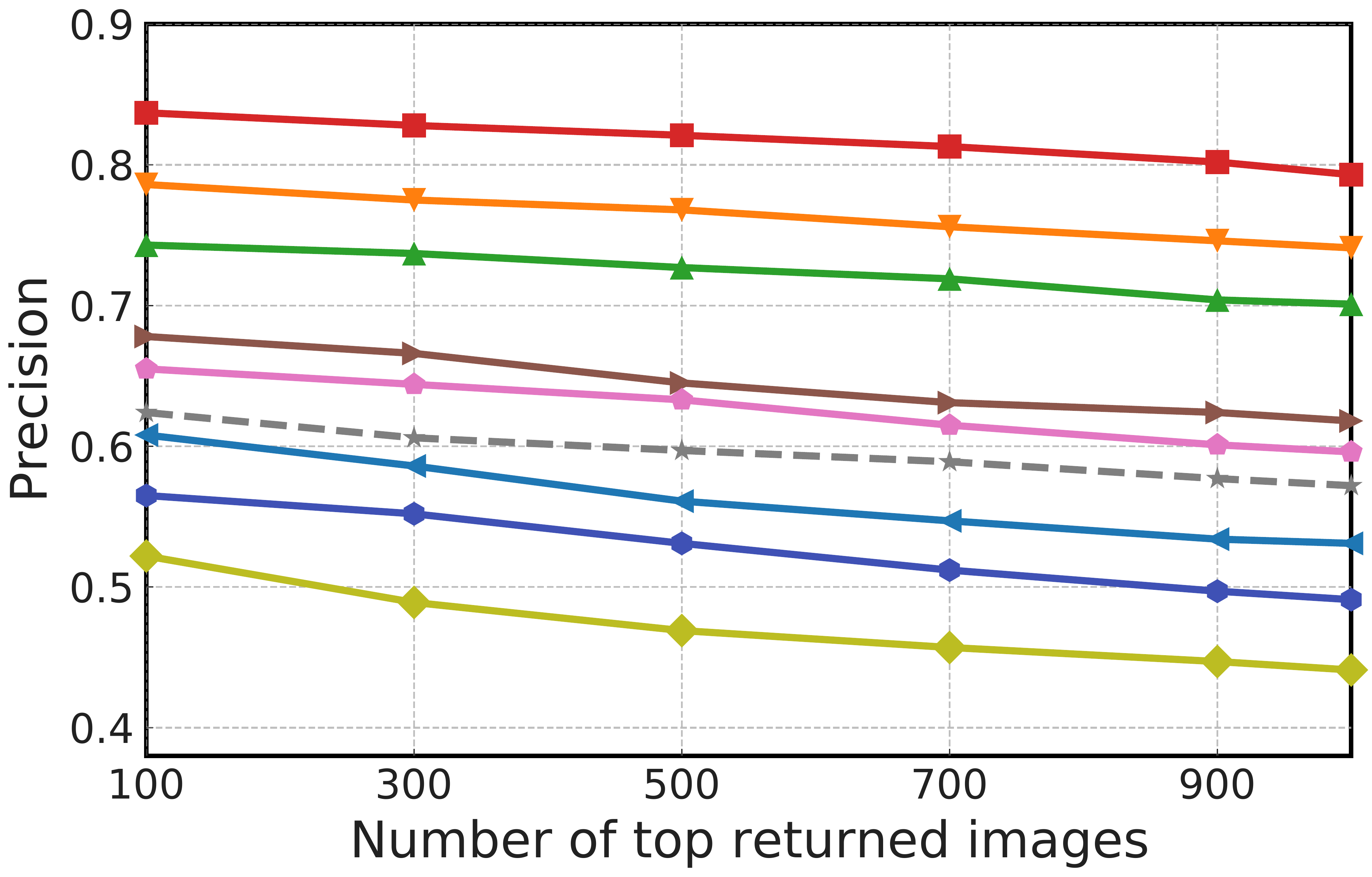}}
    \hfill
    \subfloat[\footnotesize  P@N=5000 at NUS-WIDE:64bits]{\includegraphics[width=0.49\linewidth]{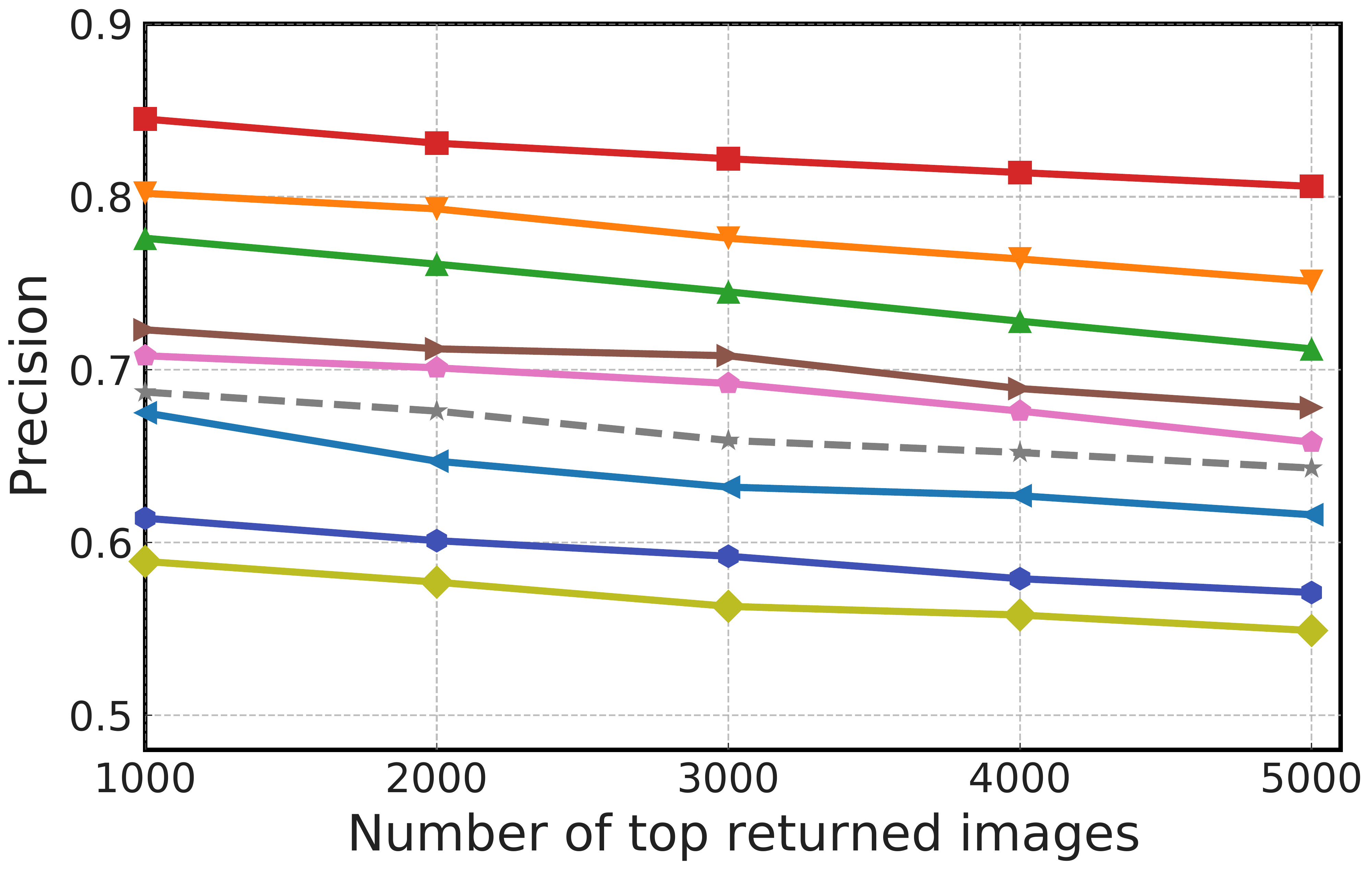}}
    \\
  \subfloat[\footnotesize  P@H$\leq$2 on ImageNet]{\includegraphics[width=0.49\linewidth]{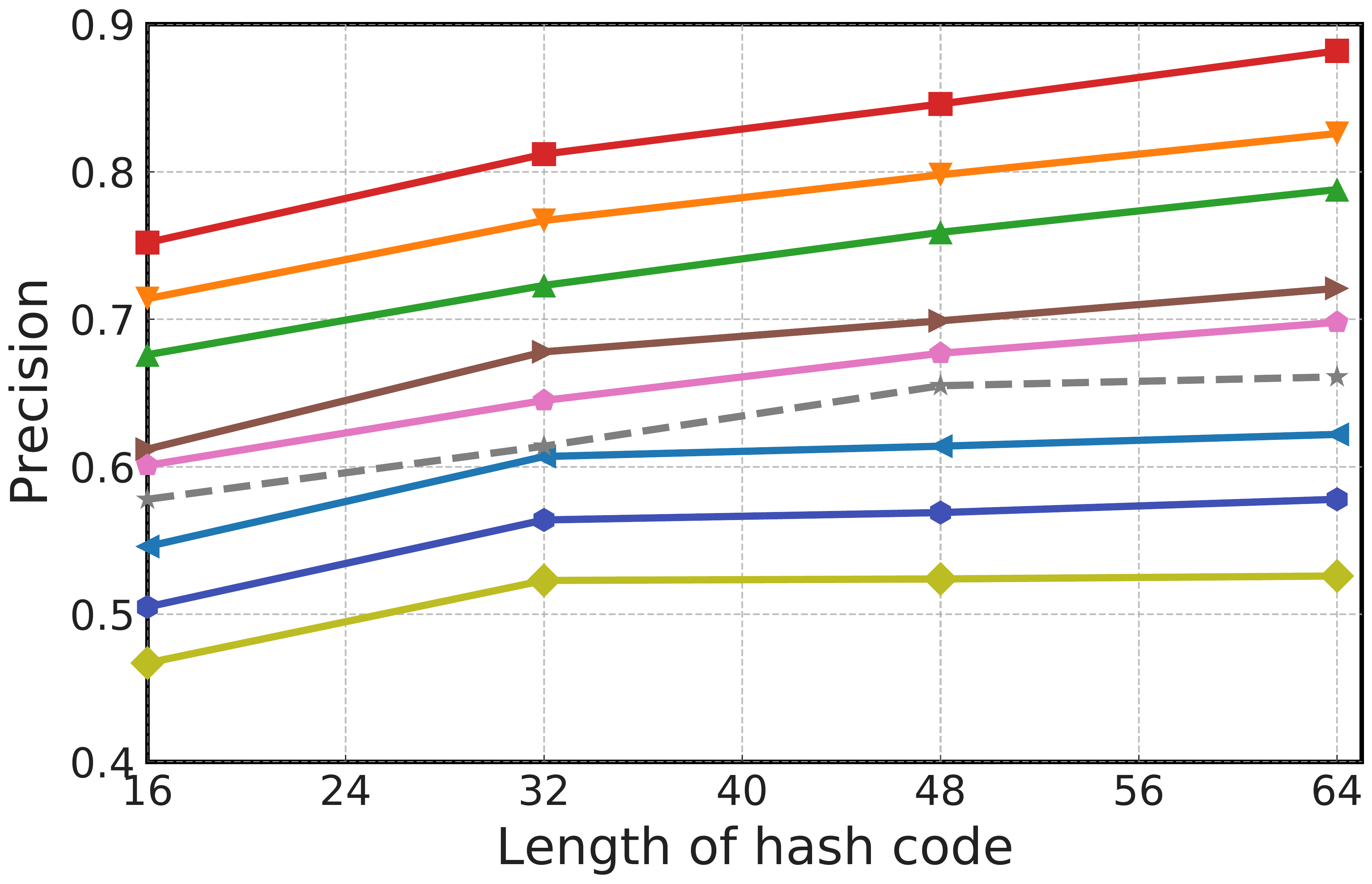}}
    \hfill
    \subfloat[\footnotesize P@H$\leq$2 on NUS-WIDE]{\includegraphics[width=0.49\linewidth]{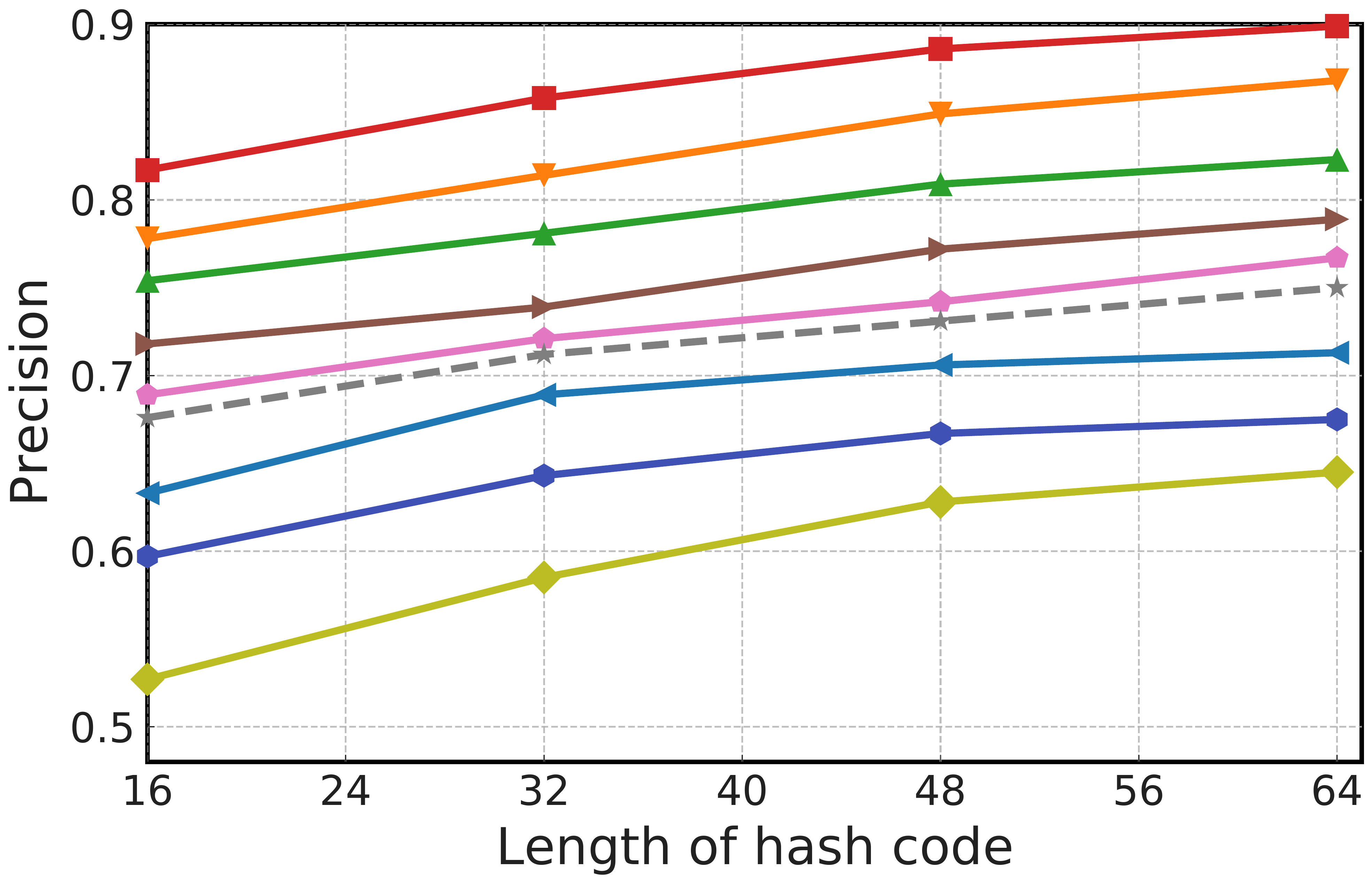}}
    \caption{ P-R curves, P@N,  and P@H$\leq$2 of \ourmethod and comparison methods on ImageNet and NUS-WIDE.}
    \label{fig:PR}
\end{figure}

\subsection{Implementation Details}
\noindent\textbf{Model details.} For fair comparisons, we follow DSCH~\cite{DSCH2022AAAI} to adopt VGG19~\cite{vgg2014ICLR} pre-trained on ImageNet~\cite{ImageNet2009CVPR} as the backbone, and use the hash layer consisting of two fully-connected layers with $ReLU$ as the activation function for hash code projection~\cite{DSCH2022AAAI}. The $3\times224\times224$ image will be transformed to a $4096-d$ feature vector and then to the $K$-bit continuous hash code. In addition, we have an auxiliary projection head $Exp\_map$ parameterized by $\theta_{e}$ after the hash layer. The $K$-bit hash code will finally be projected to a 128-bit hyperbolic embedding in the Poincar\'e ball.

\noindent\textbf{Training details.} We implement \ourmethod in PyTorch~\cite{pytorch2019NPIS} and train the model with an NVIDIA RTX 3090 GPU. Following~\cite{CIBHash2021IJCAI,DVB2019IJCV,TBH2020CVPR}, we freeze the backbone and only train the hash layer and the projection head. For data augmentation, we use the same strategy as CIBHash~\cite{CIBHash2021IJCAI} and DSCH~\cite{DSCH2022AAAI}. We set the curvature parameter $c=0.1$ for ImageNet and $c=0.01$ for other datasets (See $\S$~\ref{sec:sensitivity}). The temperature parameter $\tau=0.2$ (Equations (\ref{eqn:contrastive_instanct}) and (\ref{eqn:contrastive_proto})). The default $M$ and $L$ are set to $[1500\rightarrow1000\rightarrow800]$ for ImageNet,  $[100\rightarrow80\rightarrow50]$ for CIFAR-10, $[200\rightarrow150\rightarrow80]$ for FLICKR25K, and $[200\rightarrow150\rightarrow80]$ for NUS-WIDE (See $\S$~\ref{sec:sensitivity} for detailed investigation with different clustering settings). We set the batch size $B=64$ and adopt the Adam optimizer~\cite{Adam2015ICLR} with a learning rate $lr=0.001$.

\subsection{Comparison and Analysis}
The mAP results on four benchmark datasets are shown in Table~\ref{table:all_map}. It is clear that our proposed \ourmethod consistently achieves the best retrieval performance among the four image datasets, with an average increase of $4.5\%$, $1.6\%$, $1.3\%$, and $3.5\%$ on ImageNet, CIFAR-10, FLICKR25K, and NUS-WIDE compared with DSCH, respectively. We also report the mean intra-class distance $d_{intra}$ and mean inter-class  distance $d_{inter}$ on ImageNet in Table~\ref{tab:distance_comparison}. The results demonstrate that \ourmethod can learn more compact hash codes with greater disentangling ability than others. In addition, we report the P-R curves, P@N curves, and P@H$\leq 2$ curves at 64 bits in Figure~\ref{fig:PR}. Obviously, \ourmethod outperforms all compared methods by large margins on both ImageNet and FLICKR25K w.r.t. the three metrics. These comparisons imply that \ourmethod can generate high-quality hash codes, leading to stable superior retrieval performance.

\subsection{Ablation Study}
To justify how each component of \ourmethod contributes to final retrieval performance, we conduct studies on the effectiveness of 1) the embedding of hash codes into hyperbolic space and 2) the utilization of hierarchical semantic structures.

\noindent\textbf{Effect of hyperbolic embedding.}
We report the mAP performance of embedding in hyper-sphere space and hyperbolic space in Table~\ref{tab:distance_comparison}. In hyper-sphere space, we disable the projection head, perform hierarchical K-Means directly on the continuous hash codes, and use Equation (\ref{eqn:cos_dist}) as the distance metric to compute the contrastive loss. We observe that hyperbolic embedding can boost an average increase of $2.7\%$ and $1.6\%$ on ImageNet and FLICKR25K, respectively, which implies that hyperbolic space has superior expression ability with less distortion than hyper-sphere space.

\noindent\textbf{Effect of hierarchical semantic structures.} Table~\ref{tab:ablation_hierarchies} reports the mAP results under different settings of hierarchical semantic structures. IC and PC denote the baseline models using instance-wise contrastive learning and prototype-wise contrastive learning, respectively. They have no perception of latent hierarchical semantic structures, resulting in sub-optimal retrieval performance. Comparing the third and first row of Table~\ref{tab:ablation_hierarchies}, we can observe respective $3.4\%$ and $1.5\%$ performance gains on ImageNet and NUS-WIDE after adding the hierarchical information to the instance-wise contrastive learning. This result verifies that hierarchies can effectively help instance-wise contrastive learning to sample more accurate negative samples and mine more accurate cross-sample similarity. In addition, comparing the fourth and second row of Table~\ref{tab:ablation_hierarchies}, we can achieve respective $4.2\%$ and $2.0\%$ performance improvements on ImageNet and NUS-WIDE when employing prototype-wise contrastive learning with hierarchies. It demonstrates that accurate cross-layer affiliation provided by hierarchical semantic structures is beneficial to contrastive hashing. Finally, HIC+HPC achieves the best performance, demonstrating that \ourmethod consisting of both hierarchical instance-wise contrastive learning and hierarchical prototype-wise contrastive learning promises the full utilization of the hierarchical information from both local and global perspectives.

\begin{table}[t]
    \centering
    \caption{The mAP performance in different embedding space.}
    \begin{tabular}{ccccccc}
    \toprule
    \multirow{2}{*}{Space}&
    \multicolumn{3}{c}{ImageNet}&
    \multicolumn{3}{c}{NUS-WIDE}\\
    \cmidrule(lr){2-4} \cmidrule(lr){5-7}
    &16-bit&32-bit&64-bit&16-bit&32-bit&64-bit\\
    \midrule
        Hyper-sphere & 0.769&0.791& 0.798 & 0.790&0.806&  0.811  \\
        Hyperbolic & \bf 0.783&\bf 0.814&\bf 0.826 & \bf 0.797&\bf  0.820& \bf 0.828 \\
    \bottomrule
    \end{tabular}
    
    \label{tab:embedding_space}
\end{table}

\begin{table}[t]
    \centering
    \caption{Ablation studies on hierarchical semantic structures. IC: instance-wise contrastive learning without hierarchies; HIC: hierarchical instance-wise contrastive learning; PC: prototype-wise contrastive learning without hierarchies; HPC: hierarchical prototype-wise contrastive learning.}
    \begin{tabular}{ccccccc}
    \toprule
    \multirow{2}{*}{Setting}&
    \multicolumn{3}{c}{ImageNet}&
    \multicolumn{3}{c}{NUS-WIDE}\\
    \cmidrule(lr){2-4} \cmidrule(lr){5-7}
    &16-bit&32-bit&64-bit&16-bit&32-bit&64-bit\\
    \midrule
        IC & 0.735&0.758&0.763 &0.755&0.781&0.789\\
        PC &  0.729&0.747&0.758 &0.744&0.777&0.784\\
        HIC & 0.755&0.784&0.795&0.768&0.787&0.805 \\
        HPC & 0.750&0.782&0.798&0.761&0.788&0.802\\
    \midrule
        HIC+HPC&\bf 0.783 & \bf 0.814 & \bf 0.826&\bf 0.797&\bf  0.820& \bf 0.828\\
    \bottomrule
    \end{tabular}
    \label{tab:ablation_hierarchies}
\end{table}

\subsection{Sensitivity Analysis}
\label{sec:sensitivity}
In this section, we give a detailed analysis of the hyper-parameters in the model training phase, including the $M_{l}$ and $L$ of the hierarchical hyperbolic K-Means, the curvature parameter $c$ of the hyperbolic space, and the trade-off parameter $\lambda$. Since parameters like the batch size $B$ and the temperature parameter $\tau$, \etc, have been analyzed in the related works~\cite{SimCLR2020ICML,CIBHash2021IJCAI}, we do not experiment on these parameters.

\noindent\textbf{Sensitivity to $M_{l}$ and $L$.}
We test the model's performance with a variation of the number of layers and the number of prototypes at each layer. As shown in Table~\ref{tab:sensitive_hierarchies}, we can draw the following conclusions: 1) Deeper hierarchies can improve the retrieval performance. Compared with learning with only one layer, \ourmethod achieves $2.5\%$ and $4.3\%$ improvement on ImageNet and NUS-WIDE under the best settings, respectively. Nevertheless, the depth of the hierarchy is not linearly related to performance. There exists a trade-off between performance and computation overhead. 2) \ourmethod relies on sufficient prototypes to fully capture the latent distribution. In the setting of a three-level hierarchy, more prototypes bring more performance improvements. Similarly, excessive prototypes do not improve performance.

\begin{table}[t]
    \centering
    \caption{Sensitivity analysis on the number of layers $L$ and the number of prototypes at different layers $M_{l}$. We present the configuration of $M_{l}$ and $L$ as $M_{1}\rightarrow M_{2}\rightarrow..., \rightarrow M_{L}$.}
    \begin{tabular}{clcc}
    \toprule
    Dataset&
    Configuration of $M_{l}$ and $L$&
    mAP@64-bit\\
    \midrule
   
   \multirow{7}{*}{ImageNet}&500&0.789\\
   &1500&0.806\\
   &$1500\rightarrow1000$&0.817\\
   &$2000\rightarrow1500\rightarrow800$& 0.825\\
   &$1500\rightarrow1000\rightarrow800$&\bf 0.826\\
    &$1000\rightarrow800\rightarrow500$& 0.822\\
   &$1500\rightarrow1000\rightarrow800\rightarrow500$ &0.824\\
    \midrule
    \multirow{7}{*}{NUS-WIDE}
     &100&0.798\\
    &200&0.794\\
    &$200\rightarrow150$&0.817\\
     &$300\rightarrow120\rightarrow100$&0.826\\
      &$200\rightarrow150\rightarrow80$& \bf 0.828\\
       &$100\rightarrow80\rightarrow50$&0.822\\
        &$200\rightarrow150\rightarrow80\rightarrow40$&0.827\\
    \bottomrule
    \end{tabular}
    
    \label{tab:sensitive_hierarchies}
\end{table}

\begin{figure}[ht]
    \centering
    \subfloat[Curvature parameter $c$]{\includegraphics[width=0.48\linewidth]{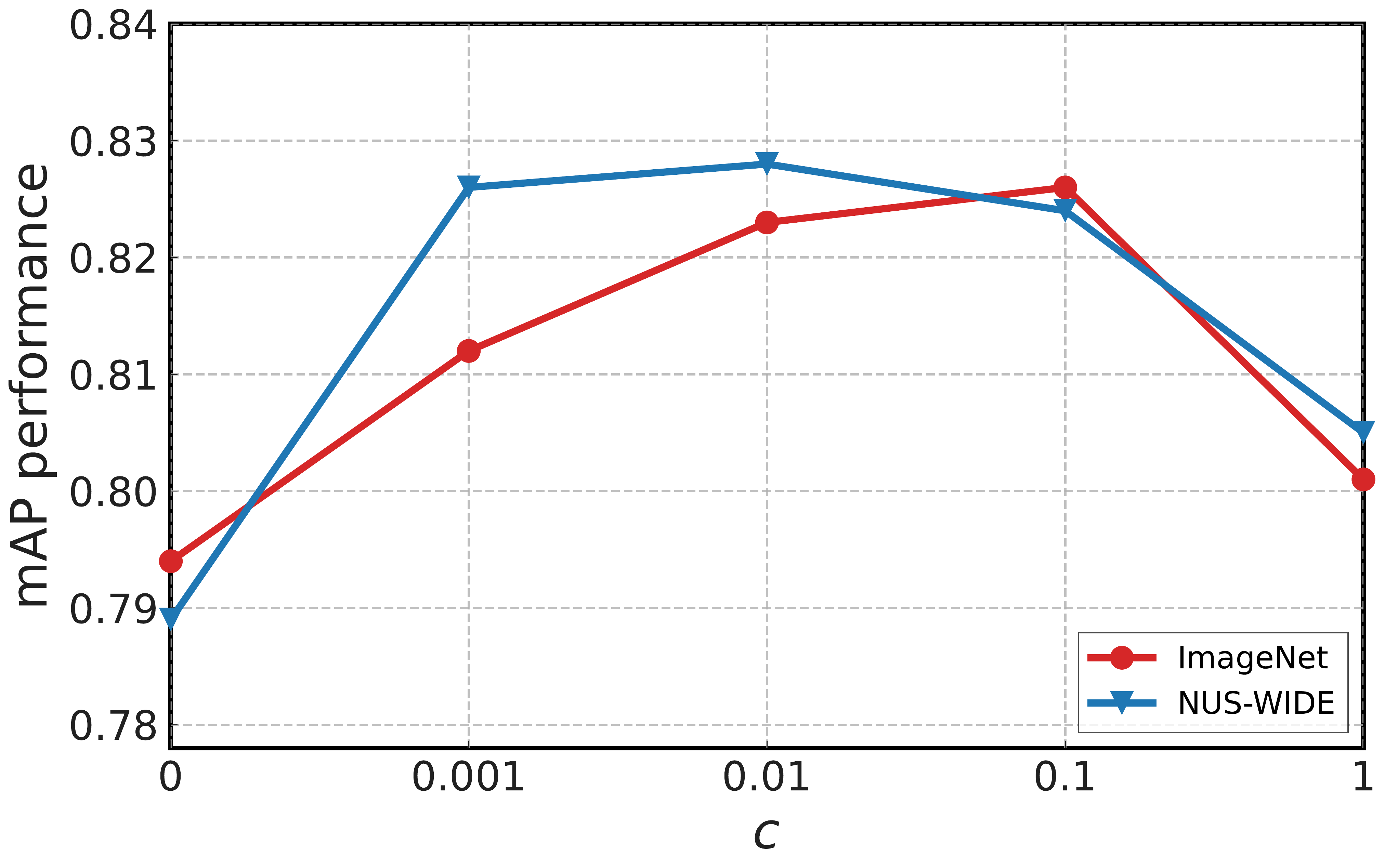}}
    \subfloat[Trade-off parameter $\lambda$]{\includegraphics[width=0.48\linewidth]{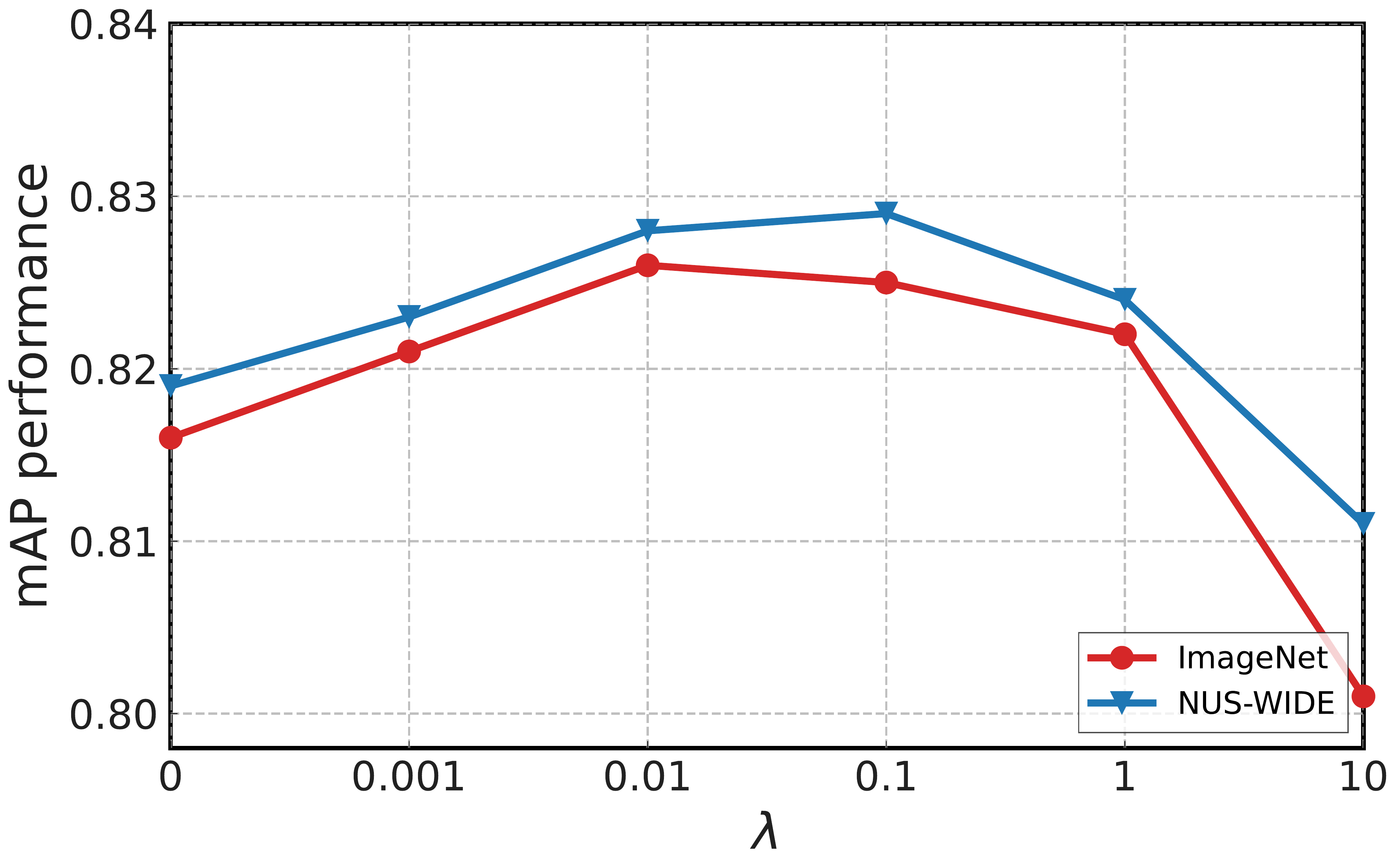}}
    \caption{The mAP performance w.r.t. different $c$ and $\lambda$ at 64 bits on ImageNet and NUS-WIDE.}
    \label{fig:sensitive_parameter}
    \vspace{-0.3cm}
\end{figure}

\noindent\textbf{Sensitivity to $c$.} We investigate the effect of the curvature parameter $c$. Intuitively, the smaller $c$ is,  the flatter the Poincar\'e ball is. As shown in Figure~\ref{fig:sensitive_parameter} (a), mAP increases as $c$ at the beginning. It gets a peak value at $c=0.01$ or $c=0.1$ but drops off sharply after $c=0.1$.
In addition, we find that the optimal mAP for ImageNet is higher than that for NUS-WDIE. We attribute it to the clear hierarchical semantic structures of the ImageNet dataset that are organized according to the WordNet~\cite{WordNet2010} hierarchy.

\noindent\textbf{Sensitivity to $\lambda$.} We test the model's performance depending on the trade-off parameter $\lambda$ at 64 bits on ImageNet and NUS-WIDE. As shown in Figure~\ref{fig:sensitive_parameter} (b), we can observe that both small and large $\lambda$ will decrease the mAP performance. A small $\lambda$ can not reduce the accumulated quantization error caused by the continuous relaxation, resulting in considerable information loss. In contrast, a large $\lambda$ will force the quantization loss item to dominate the overall learning objective, resulting in the difficulty of optimization. As a result, we opt for $\lambda=0.01$.

\subsection{Visualization}

\begin{figure}[t]
    \centering
    \subfloat[\scriptsize CIFAR-10:DSCH]{\includegraphics[width=0.24\linewidth]{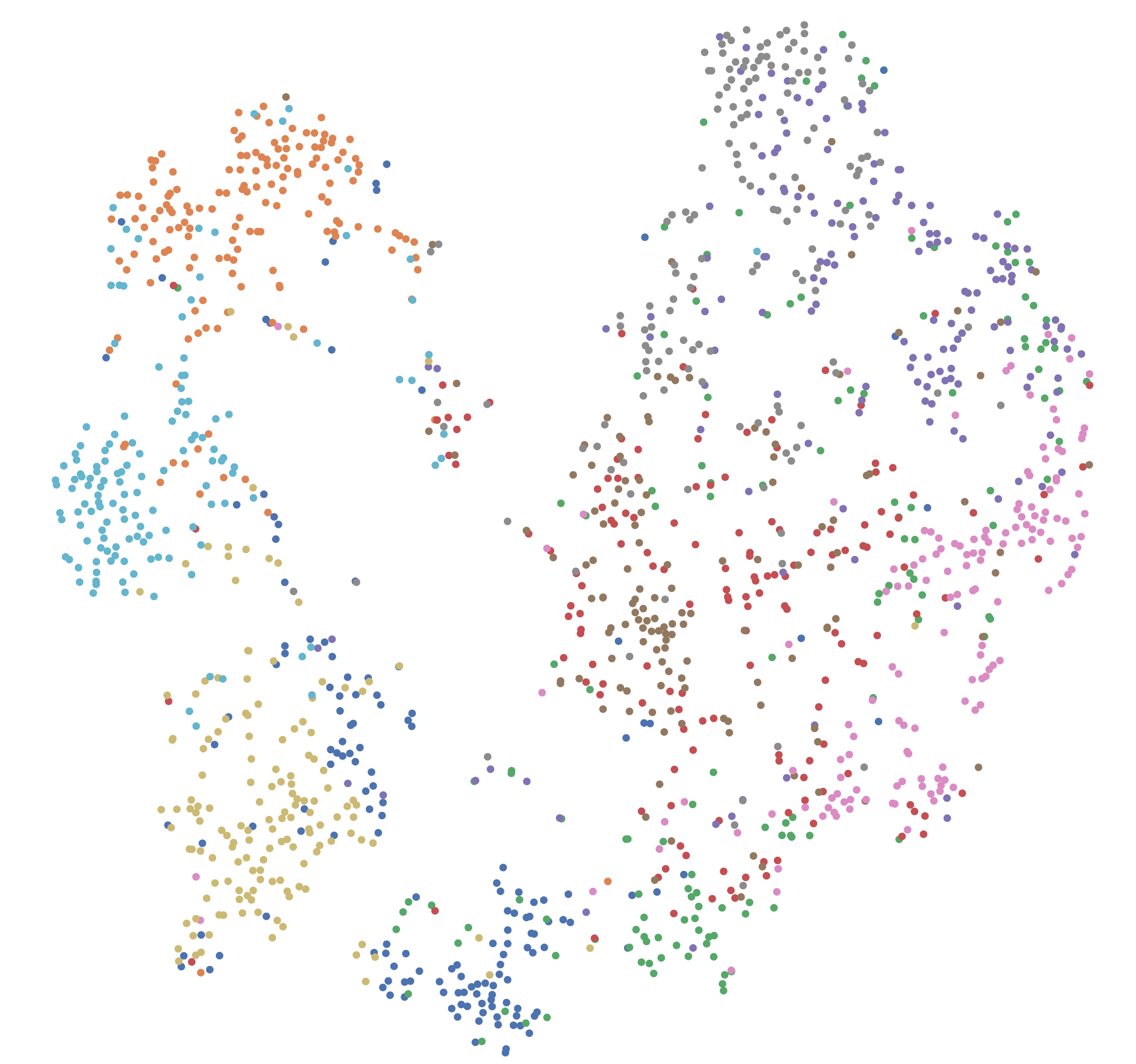}}
    \subfloat[\scriptsize CIFAR-10:\ourmethod]{\includegraphics[width=0.24\linewidth]{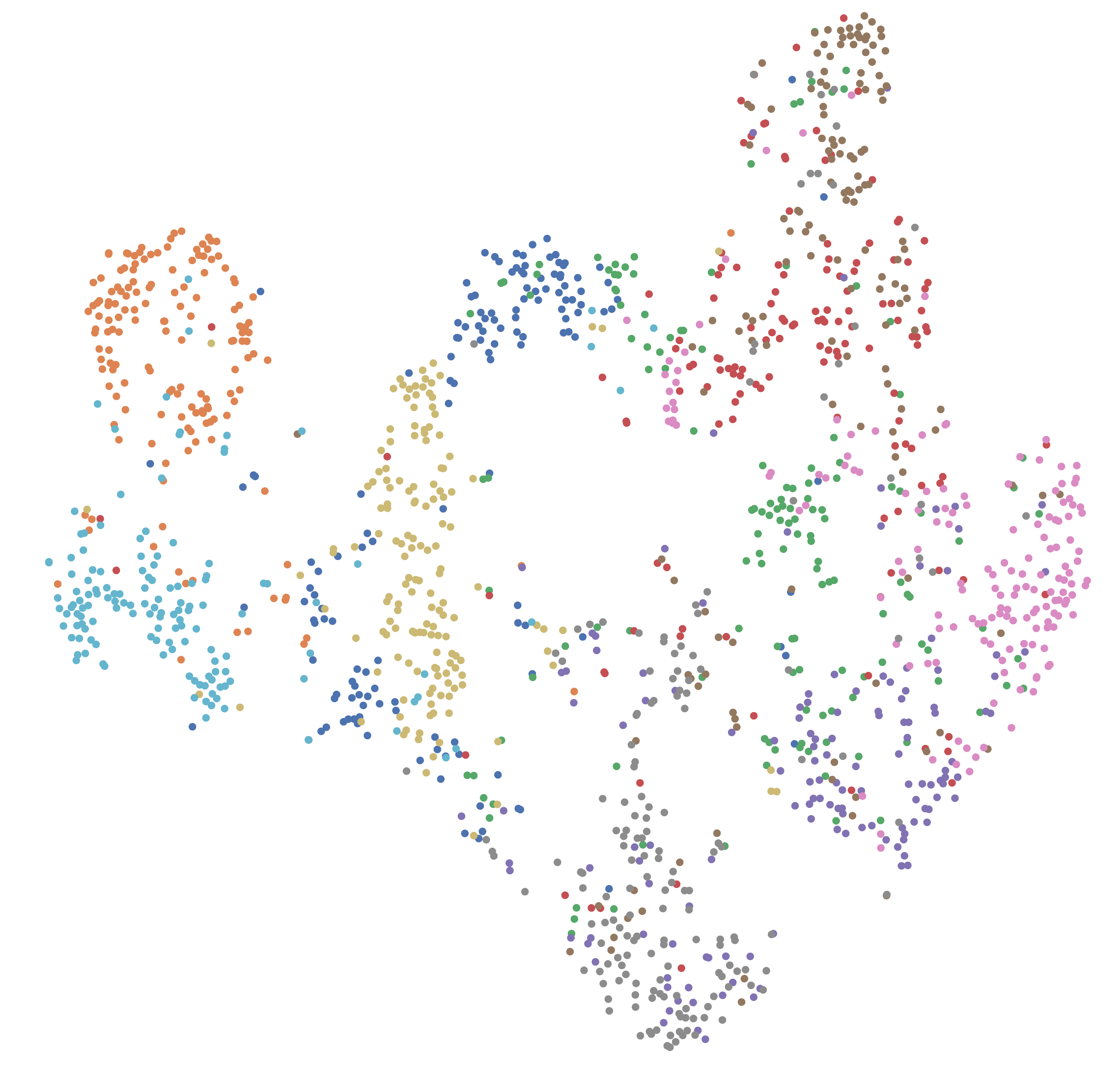}}
    \subfloat[\scriptsize ImageNet:DSCH]{\includegraphics[width=0.24\linewidth]{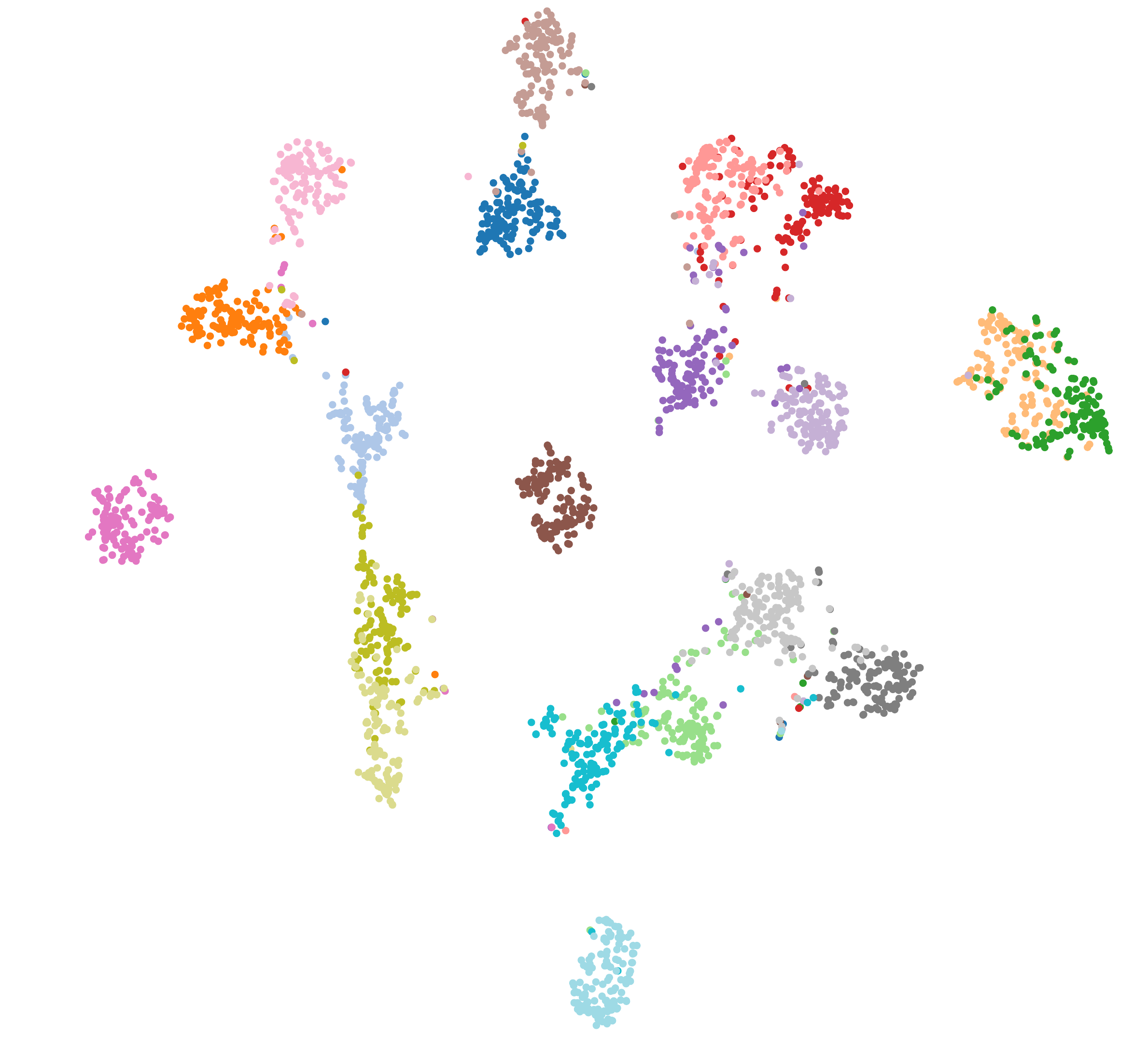}}
    \subfloat[\scriptsize ImageNet:\ourmethod]{\includegraphics[width=0.24\linewidth]{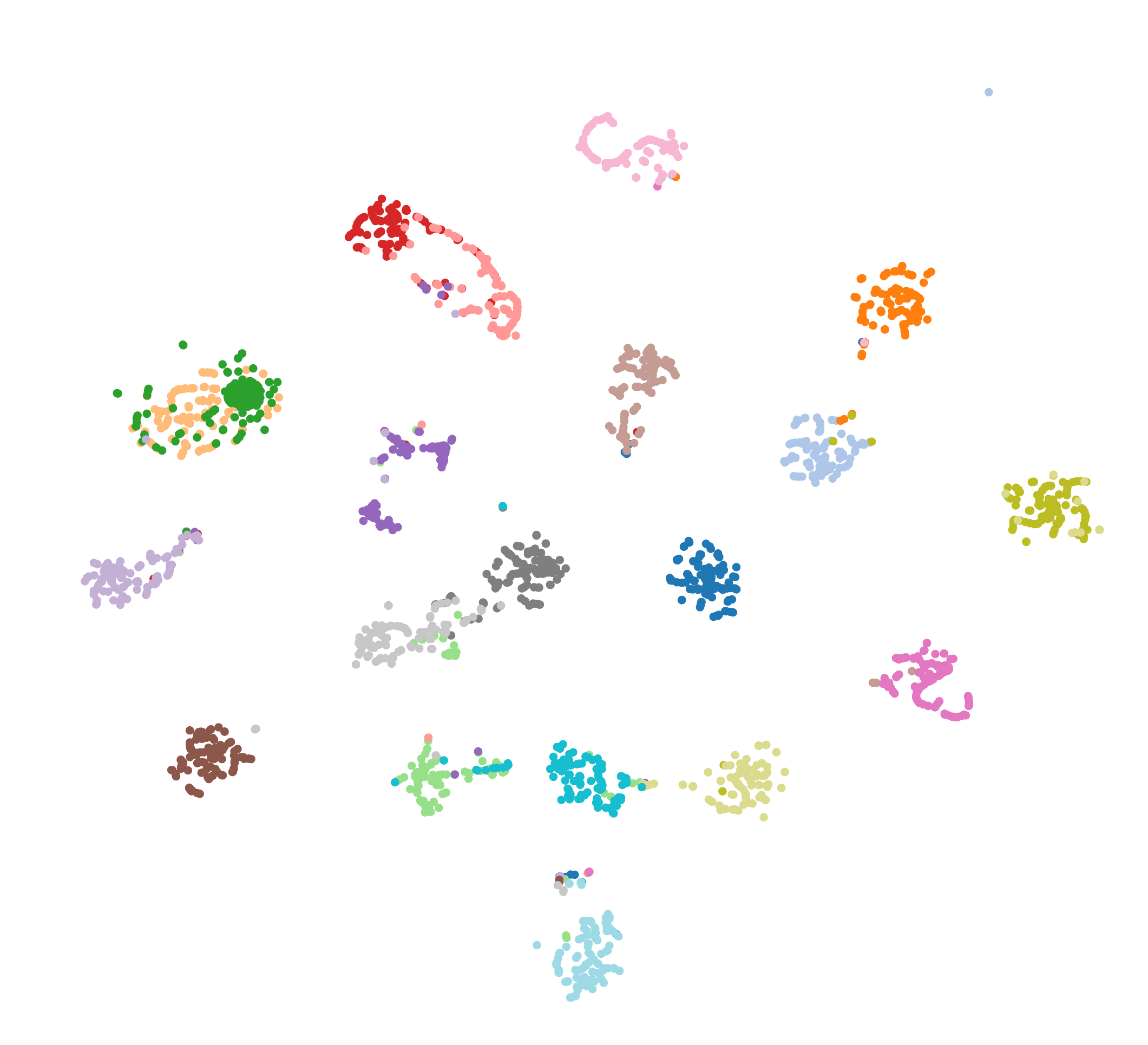}}
    \caption{The t-SNE visualization of the learned 64-bit hash codes from the training sets. The scattered elements of the same color indicate the same category. Note that we only visualize the first 20 classes for ImageNet.}
    \label{fig:t-sne}
    \vspace{-0.5cm}
\end{figure}

\noindent\textbf{Hash Codes Visualization.} Figure~\ref{fig:t-sne} shows the t-SNE visualization~\cite{t-sne2008JMLR} of the hash codes at 64 bits on CIFAR-10 and ImageNet. The hash codes generated by \ourmethod show favorable intra-class compactness and inter-class separability compared with the state-of-the-art hashing method DSCH. It demonstrates that \ourmethod can generate high-quality hash codes.

\noindent\textbf{Hyperbolic Embeddings Visualization.} We illustrate the hyperbolic embeddings of CIFAR-10 and ImageNet in the Poincar\'e ball using UMAP~\cite{UMAP2018} with the ``hyperboloid" distance metric~\cite{HyperViT2022CVPR}. We can see that the samples are clustered according to the labels, and each cluster is pushed to the border of the ball, indicating that the learned embeddings are distinguishable enough.

\begin{figure}[t]
    \centering
    \subfloat[CIFAR-10]{\includegraphics[width=0.48\linewidth]{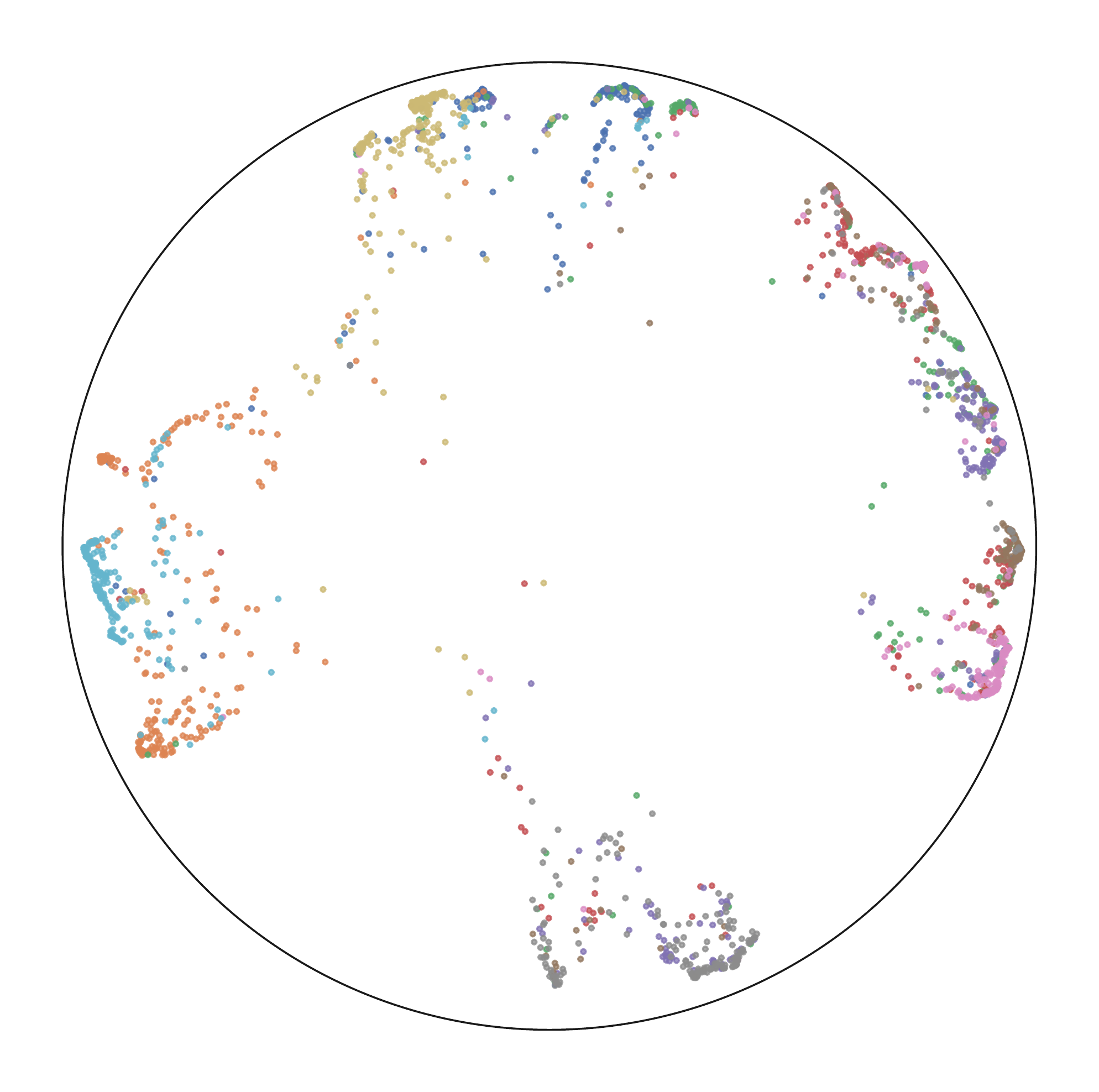}}
    \subfloat[ImageNet]{\includegraphics[width=0.48\linewidth]{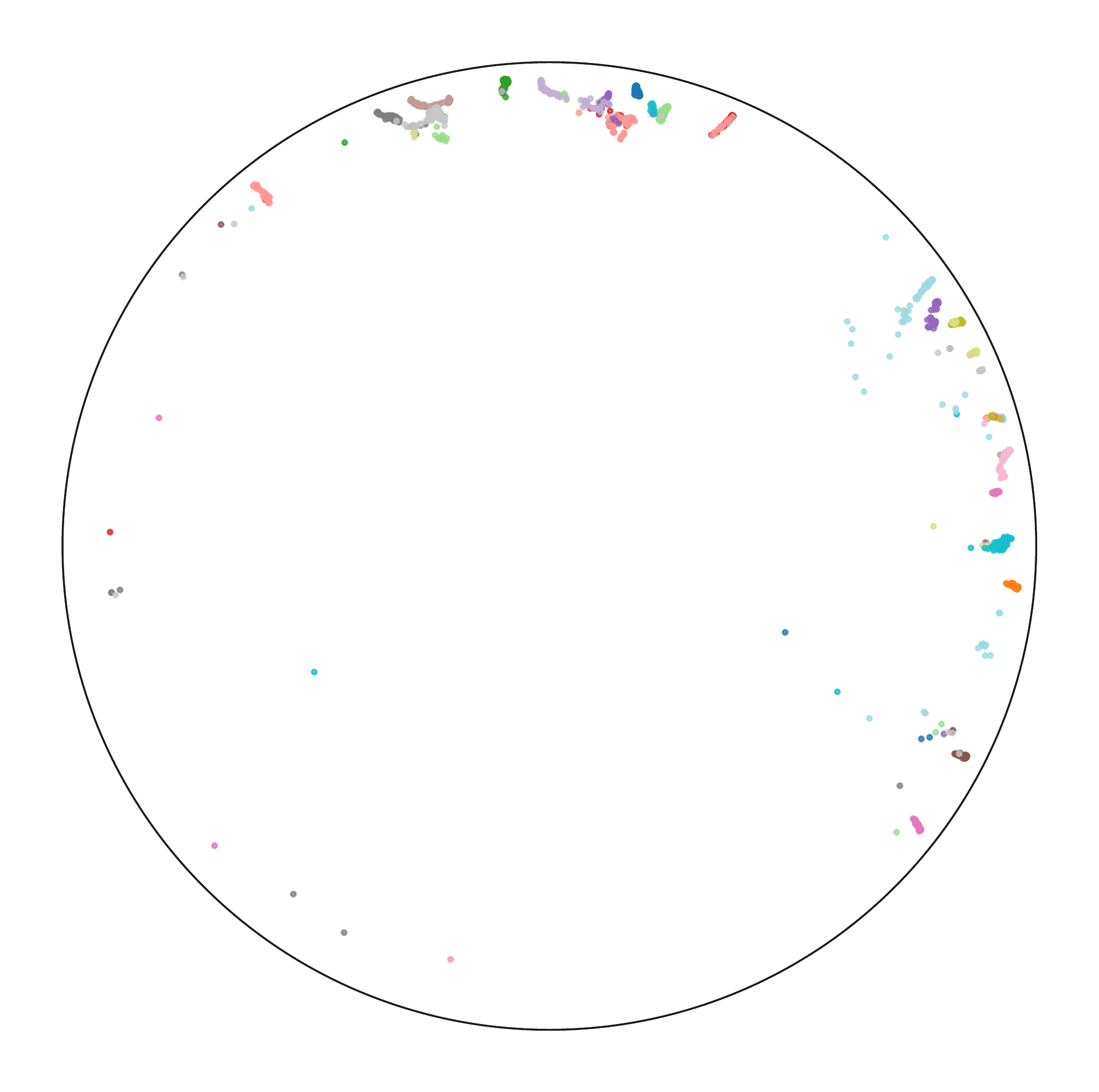}}
    \caption{The UMAP visualization of 128-d hyperbolic embeddings from ImageNet (The first 20 classes) and CIFAR-10 in the Poincar\'e ball.}
    \label{fig:hyper_visualization}
\end{figure}

\section{Conclusion}
In this paper, we propose to learn hash codes by exploiting the hierarchical semantic structures that naturally exist in real-world datasets. As a result, we proposed a novel unsupervised hashing method named \ourmethod. In \ourmethod, we embed the continuous hash codes into hyperbolic space (\ie, the Poincar\'e ball) to achieve less information distortion. Furthermore, we extend the K-Means algorithm to hyperbolic space and perform hierarchical hyperbolic K-Means to capture the latent hierarchical semantic structures adaptively. In addition, we designed hierarchical contrastive learning, including hierarchical instance-wise contrastive learning and hierarchical prototype-wise contrastive learning, to take full advantage of the hierarchies. Extensive experiments on four benchmarks demonstrate that \ourmethod can benefit from the hierarchies and outperforms the state-of-the-art unsupervised methods.

\newpage

\bibliographystyle{ACM-Reference-Format}
\bibliography{reference}

\newpage

\appendix

\section*{Appendix}

\section{More details about the framework}
Figure~\ref{fig:framework} shows the framework of \ourmethod. Following SimCLR~\cite{SimCLR2020ICML}, we adopt Siamese networks directly sharing parameters as our learning framework. We first augment the training images with a combination of different augmentation strategies to generate various views of the same image. The augmentation strategies include random crop, color jittering, Gaussian blur, \textit{etc}.  We can feed the $224\times224\times3$-d images to the VGG19~\cite{vgg2014ICLR} backbone, and obtain the $4096$-d feature vectors. Then, we employ the hash layer, including two fully-connected layers with the ReLU function as the activation function, to transform them into $K$-d features. Furthermore, all these features will be constrained to $(-1,1)$ via the $tanh$ function for continuous relaxation. After that, the projection head, consisting of a fully connected layer and the exponential mapping function, will map all the continuous hash codes into the hyperbolic space (\ie, the Poincar\'e ball) with the hyperbolic embedding dimension of 128, where we perform hierarchical contrastive learning with the captured hierarchical semantic structures. 

In the test phase, we disable the projection head and generate hash codes with the VGG backbone and the well-trained hash layer. We use the modified $sgn$ function for binarization:

\begin{equation}
    \begin{aligned}
     sgn(h_{k})=\begin{cases}
 -1, &\text{if } h_{k}\le 0,\\
 1, &otherwise,
\end{cases}
    \end{aligned}
\end{equation}
where $h_{k}$ is the $k$-th bit of the hash code $h$.

\section{More details about datasets}

The details of the dataset setting can be found in Table~\ref{tab:datasets}. For ImageNet and NUS-WIDE, we follow~\cite{CSQ2020CVPR,DCH2018CVPR} to use the commonly adopted index files and compressed datasets from the repository of HashNet~\cite{HashNet2017CVPR} to form the splits. Besides, we follow~\cite{BitEntropy2021AAAI} to implement CIFAR-10 and FLICKR25K.

\begin{table}[h]
    \centering
	\caption{Experimental settings for all datasets.}
	\label{tab:datasets}
	\scalebox{1.1}
	{
		\begin{tabular}{lccccc} 
			\toprule
			Dataset    & \#Train & \#Query & \#Retrieval &\#Class  \\ \hline
		    ImageNet  & 1,3000 & 5,000 & 128,503 & 100\\
		    CIFAR-10  & 10,000  & 1,000  & 59,000  & 10   \\
			FLICKR25K  & 10,000  & 1,000  & 24,000  & 24   \\
			NUS-WIDE     & 10,500  & 2,100  & 193,734    & 21  \\
			\bottomrule
	\end{tabular}
	}
	\vspace{-0.55cm}
\end{table}

\section{Top-10 Retrieved Results} 

\begin{figure*}[t]
    \centering
    \includegraphics[width=\textwidth]{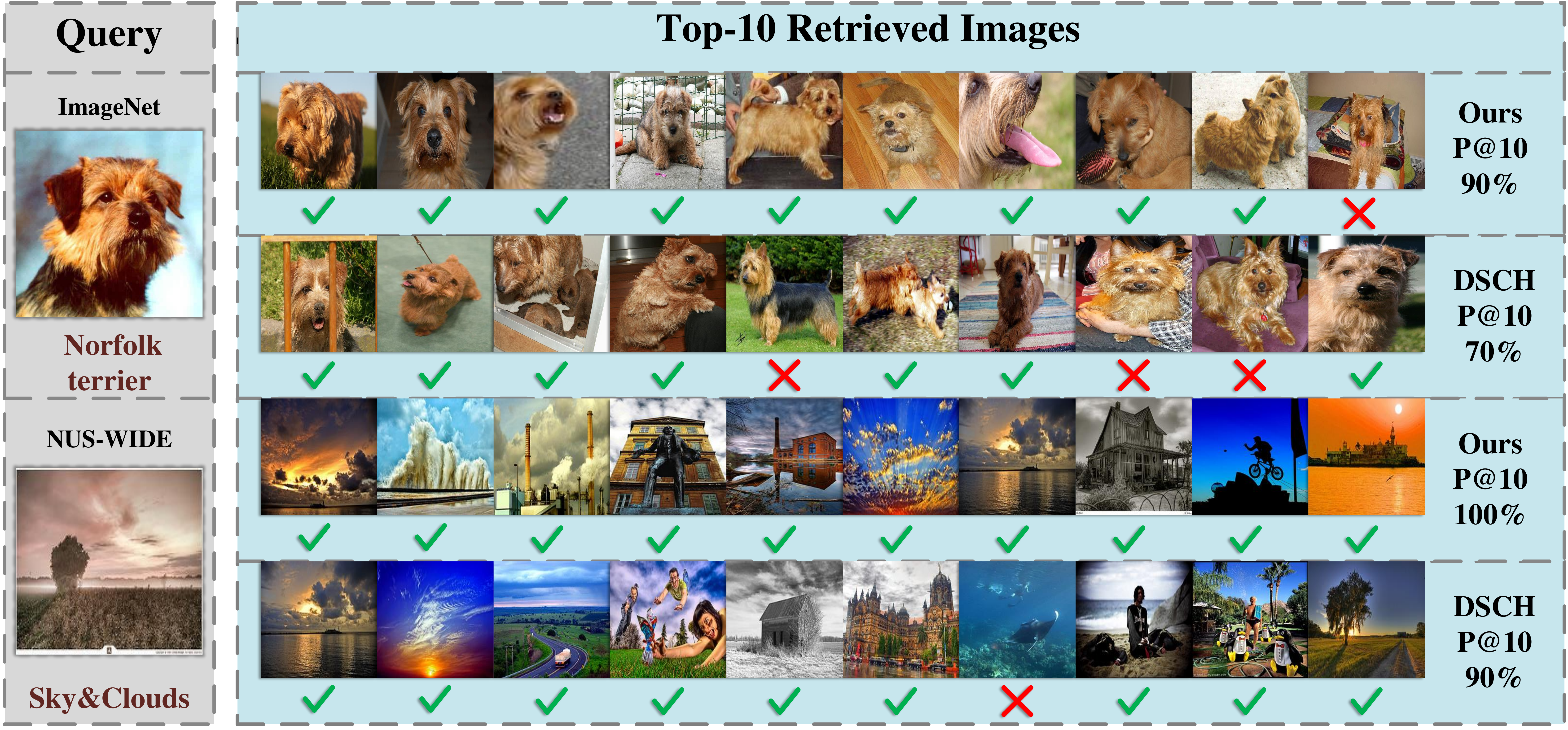}
    \caption{Retrieval comparisons to DSCH~\cite{DSCH2022AAAI} at 64 bits on three datasets. The Top-10 retrieved images are returned according to the Hamming distance between the query image and the database images. We report precision within the top-10 retrieved images (P@10) and the results demonstrate that our proposed \ourmethod outperforms DSCH with more relevant and accurate returned images as well as fewer contradictions.}
    \label{fig:query}
\end{figure*}

Figure~\ref{fig:query} illustrates the top-10 retrieved images and reports P@10 comparisons between \ourmethod and DSCH~\cite{DSCH2022AAAI}. Our proposed \ourmethod achieves $90\%$ and $100\%$ in terms of P@10 when given the query images labeled as ``Norfolk terrier'' and ``Sky\&Clouds''  on ImageNet and NUS-WIDE, respectively. It can be seen that our proposed \ourmethod yields more relevant and accurate retrieval results than DSCH.

\section{Visualization of Hierarchical Semantics}
\begin{figure}[htb]
    \centering
    \includegraphics[width=\linewidth]{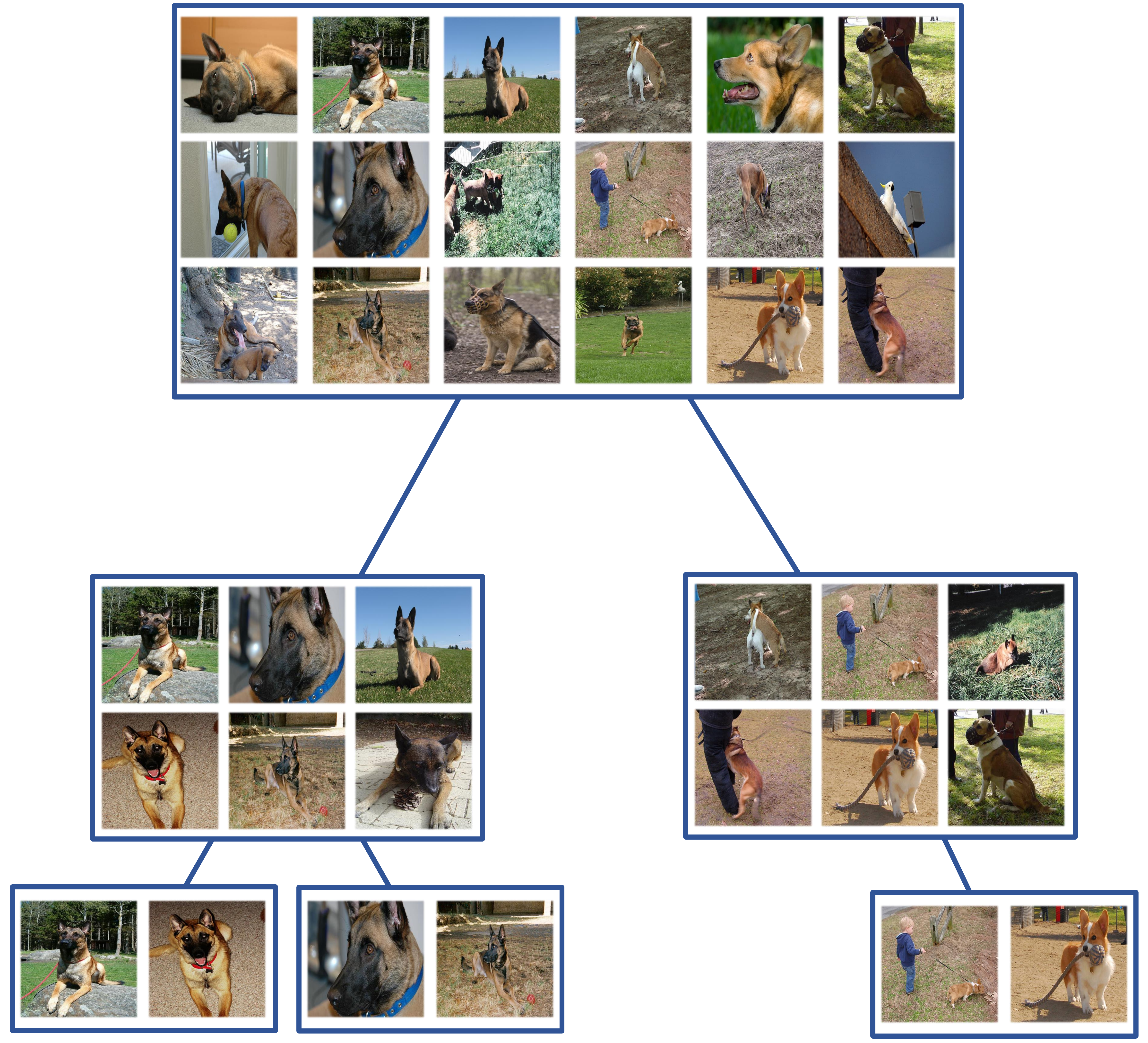}
     \caption{Visualization of a captured hierarchical semantic structure. Note that we only use a small part of the images for the visualization.}
    \label{fig:hierarchies_visualization}
\end{figure}

In Figure~\ref{fig:hierarchies_visualization}, we visualize the partial results of the hierarchical hyperbolic K-Means. It is clear that images at low layers express finer-grained semantics and the high layers contain coarser-grained semantics, \eg, images at the bottom of the hierarchy are naturally more visually similar, while the images at the top of the hierarchy are more diverse. These results indicate that \ourmethod is capable of capturing the hierarchical semantics of the data very well.

\section{Time Complexity Analysis}
Since \ourmethod involves an extra hierarchical hyperbolic K-Means algorithm before each epoch for the training phase, we discuss the possible extra time overhead according to time complexity. Note that $N$, $M_{l}$, $L$, and $B$ denote the dataset size, number of prototypes at the $l$-th layer, number of layers, and the mini-batch size, respectively.

On the one hand, the time complexity of vanilla K-Means is $\mathcal{O}(NMt)$, where $t$ is the number of iterations in K-Means and we set $t=30$.
For hierarchical hyperbolic K-Means, the extra time complexity of each training step is $\mathcal{O}(NM_{1}t+M_{1}M_{2}t+...+M_{L-1}M_{L}t)/(N/B)$. Since $M_{l}\ll N$, we can simplify it to $\mathcal{O}(BM_{1}t)$.

On the other hand, the time complexity of hierarchical instance-wise contrastive learning is $\mathcal{O}(B^2)$, and the counterpart of hierarchical prototype-wise contrastive learning is $\mathcal{O}(BM_{1}+...+BM_{L})=\mathcal{O}(BM_{1})$. As a result, the time complexity of hierarchical contrastive loss computation is $\mathcal{O}(B^2)+\mathcal{O}(BM_{1})$.

In conclusion, the complete time complexity of \ourmethod is $\mathcal{O}(B^2)+\mathcal{O}(BM_{1})+\mathcal{O}(BM_{1}t)=\mathcal{O}(BM_{1}t)$, which is consistent with state-of-the-art contrastive hashing methods DSCH~\cite{DSCH2022AAAI} but a little higher than CIBHash~\cite{CIBHash2021IJCAI} when $M_{1}t>B$.

\end{document}